\documentclass[sigconf]{acmart}

\usepackage{bm}
\usepackage{algorithm}
\usepackage[noend]{algorithmic}
\usepackage{subfigure}
\usepackage{balance}
\newcommand{\IFLINE}[2]{\STATE\algorithmicif\ #1\ \algorithmicthen\ #2}

\newtheorem{proposition}{Proposition}

\AtBeginDocument{%
  }

\copyrightyear{2026}
\acmYear{2026}
\setcopyright{cc}
\setcctype{by}
\acmConference[KDD 2026] {Proceedings of the 32nd ACM SIGKDD Conference on Knowledge Discovery and Data Mining V.2}{August 9--13, 2026}{Jeju Island, Republic of Korea.}
\acmBooktitle{Proceedings of the 32nd ACM SIGKDD Conference on Knowledge Discovery and Data Mining V.2 (KDD 2026), August 9--13, 2026, Jeju Island, Republic of Korea}
\acmISBN{979-8-4007-2259-2/2026/08}
\acmDOI{10.1145/3770855.3817736}
\begin{document}

\title{Backward Compatibility in Tree-Based Explanations and Enhanced CART Algorithm}

\author{Hirofumi Suzuki}
\email{suzuki-hirofumi@fujitsu.com}
\orcid{0000-0003-0002-9105}
\affiliation{%
  \institution{Fujitsu Limited}
  \city{Kawasaki}
  \state{Kanagawa}
  \country{Japan}
}

% \renewcommand{\shortauthors}{Hirofumi Suzuki}

\begin{abstract}
In the operation of machine learning models, \emph{model update} is a fundamental process that requires careful consideration of its impact on downstream decision-making.
Particularly when operating explainable models, changes in explanations resulting from model updates can lead to detrimental outcomes for users.
\emph{Decision trees}, due to their high transparency, are frequently employed in risk-sensitive decision-making and serve as a prominent example in which the aforementioned issue is evident.
However, existing research addressing similar issues has focused on explanations based on feature contributions, and thus cannot handle explanations derived from tree structures.
Therefore, this paper proposes the \emph{Backward Compatibility Loss in Tree-based eXplanations (BCLTX)}, a loss metric that suppresses changes in decision tree explanations before and after updates.
Furthermore, we design \emph{CART with Backward Compatibility in Tree-based eXplanations (CART-BCTX)}, a lightweight algorithm that improves upon CART for the decision tree update problem under BCLTX.
Experimental results using 10 real-world datasets, including both classification and regression tasks, show that CART-BCTX achieves favorable trade-offs between prediction performances and BCLTX values, with comparable computation times to CART, regardless of the task.
\end{abstract}

\begin{CCSXML}
<ccs2012>
<concept>
<concept_id>10010147.10010257.10010293.10003660</concept_id>
<concept_desc>Computing methodologies~Classification and regression trees</concept_desc>
<concept_significance>500</concept_significance>
</concept>
</ccs2012>
\end{CCSXML}

\ccsdesc[500]{Computing methodologies~Classification and regression trees}

\keywords{Decision Tree, Backward Compatibility, Explainable AI, CART}
\begin{teaserfigure}
    \centering
    \includegraphics[width=\textwidth]{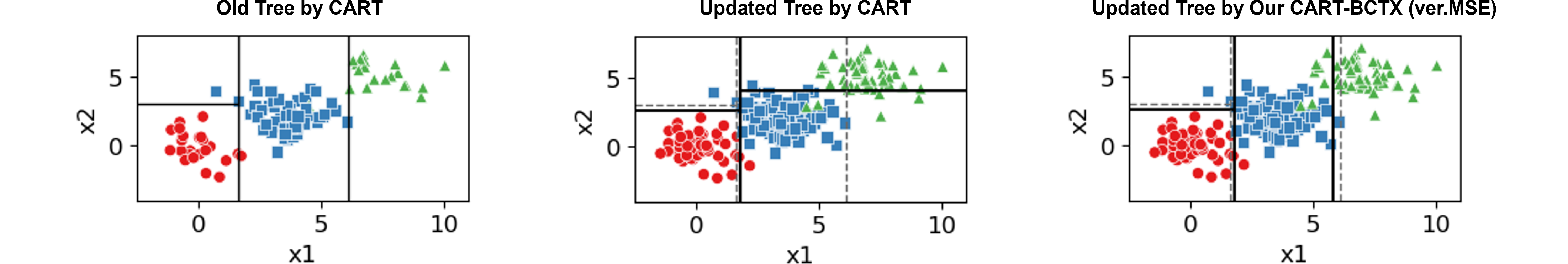}
    \caption{
        Solid and dashed lines denote the decision boundaries of the corresponding decision tree and the old decision tree, respectively.
        When a naive decision tree update is performed using CART on a new dataset, the explanations (i.e., the feature subspaces associated with predictions) change substantially for many samples belonging to the blue square and green triangle classes.
        In contrast, updates performed using our CART-BCTX result in only minor changes in explanations for most samples.
    }
    \Description{Concept of this paper.}
    \label{fig:concept}
\end{teaserfigure}


\maketitle
\newcommand\kddavailabilityurl{https://doi.org/10.6084/m9.figshare.32521632}
\ifdefempty{\kddavailabilityurl}{}{
\begingroup\small\noindent\raggedright\textbf{Resource Availability:}\\
% please change the following context to include multiple artifacts if necessary, including data, models, code, etc.
The source code of this paper has been made publicly available at \url{\kddavailabilityurl}.
\endgroup
}

%%%%%%%%%%%%%%%%%%%%%%%%%%%%%%%%%%%%%%%%%%%%%%%%%%
%%%%% Introduction
%%%%%%%%%%%%%%%%%%%%%%%%%%%%%%%%%%%%%%%%%%%%%%%%%%
\section{Introduction}
In the operation of systems based on machine learning models, it is necessary to continuously monitor model performance and conduct \emph{model updates} as needed in order to maintain or improve their performance~\cite{MLOps-Symeonidis2022,MLOps-Kreuzberger2023,MLOps-Paleyes2023}.
Model updates replace models trained on older datasets with those trained on more recent and/or larger datasets accumulated during system operation.
However, such updates inevitably change model behavior and can have a substantial impact on downstream decision-making, potentially leading to unexpected or undesirable outcomes.
Indeed, a user study has reported that naive model updates reduce the productivity of decision-making supported by machine learning models~\cite{BackwardCompatibility-Bansal2019}.
To address this issue, prior studies have investigated \emph{backward compatibility}, a metric that quantifies differences in model behavior before and after updates, as well as model update methods based on this concept~\cite{BackwardCompatibility-Bansal2019,BackwardCompatibility-Srivastava2020,BackwardCompatibility-Sakai2022,BackwardCompatibility-Matsuno2023,BackwardCompatibility-Exp-Matsuno2024,BackwardCompatibility-Exp-Yamakura2025}.

Explainable models~\cite{Explainability-Survey-Dosilovic2018,Explainability-Survey-BarredoArrieta2020} are widely deployed in risk-sensitive decision-making domains such as healthcare, finance, and manufacturing, where model updates can lead to detrimental outcomes for users.
In addition, even when an updated model achieves superior predictive performance, users may not fully trust it if its explanations differ substantially from those of the previously deployed model~\cite{Wang2023-WatchOutForUpdates}, highlighting the importance of backward compatibility in explanations.
A representative example of explainable models is \emph{decision trees}~\cite{CART-Breiman1984}, which provide explanations through if–then–else rules derived from tree structures and are frequently used in risk-sensitive applications due to their high transparency~\cite{Apps-FaultDetection-Zhao2012,Apps-FinancialFailure-Ocal2015,Apps-Diagnosis-Ghiasi2020,Apps-PredictiveMaintenance-Kaparthi2020}.
However, existing studies on backward compatibility~\cite{BackwardCompatibility-Exp-Matsuno2024,BackwardCompatibility-Exp-Yamakura2025} primarily focus on SHAP-based feature contribution explanations~\cite{Explainability-SHAP-Lundberg2017}, and thus cannot handle tree-based explanations.

To overcome this limitation, we propose the \emph{Backward Compatibility Loss in Tree-based eXplanations (BCLTX)}, a loss metric that captures changes in tree-based explanations before and after model updates.
Unlike existing backward compatibility metrics~\cite{BackwardCompatibility-Exp-Matsuno2024}, the core idea of BCLTX does not rely on the conventional notion of disagreement measures~\cite{Explainability-DisagreementProblem-Krishna2024}, but is instead specifically designed to directly reflect the properties of tree-based explanations.
Furthermore, we formulate a decision tree update problem induced by BCLTX, and propose \emph{CART with Backward Compatibility in Tree-based eXplanations (CART-BCTX)} to solve this problem.
CART-BCTX is a lightweight algorithm that extends CART~\cite{CART-Breiman1984}, a standard decision tree learning algorithm, allowing backward compatible updates to be performed efficiently without sacrificing scalability.
The main concept of this paper is shown in Figure~\ref{fig:concept}.

Our contributions are summarized as follows:
\begin{itemize}
\item
We focus on the fact that decision tree explanations can be represented as a conjunction of intervals defined by split points for individual features, and propose a backward compatibility loss metric that captures explanation differences before and after updates.
The proposed metric includes formulations based on the Jaccard coefficient, which captures differences in contributing feature sets, as well as formulations based on absolute and squared errors, which capture differences in feature-wise intervals.
\item
Based on the proposed backward compatibility loss metrics, we formulate a decision tree update problem and design an efficient algorithm to solve it.
The proposed algorithm extends the standard decision tree learning algorithm CART; compared with CART, it has equivalent computational complexity when using Jaccard- or squared-error–based loss metrics, and incurs at most a $\log n$ factor increase in the worst case when using absolute-error–based loss metrics, where $n$ denotes the number of samples.
\item
We conducted experiments on 10 real-world datasets, including both classification and regression tasks.
The results demonstrate that our update algorithm achieves excellent trade-offs between predictive performances and the proposed loss metrics, with computation times comparable to those of CART.
In addition, the results suggest that the proposed loss metrics based on absolute error and squared error contribute to improving backward compatibility with respect to predictions.
\item
Although our framework assumes a batch learning environment, backward compatibility can also be evaluated in a batch-wise manner under incremental learning with data streams. 
Based on this perspective, we conduct comparative experiments with incremental decision trees. 
The results suggest that backward compatibility in tree-based explanations is not necessarily preserved over medium- to long-term periods by incremental decision trees.
\end{itemize}

%%%%%%%%%%%%%%%%%%%%%%%%%%%%%%%%%%%%%%%%%%%%%%%%%%
%%%%% Related Work
%%%%%%%%%%%%%%%%%%%%%%%%%%%%%%%%%%%%%%%%%%%%%%%%%%
\section{Related Work}
Bansal et al. pointed out that changes in model behavior caused by model updates can reduce the productivity of human–AI collaborative decision-making, and introduced \emph{backward compatibility} as a metric to capture this issue~\cite{BackwardCompatibility-Bansal2019}.
Backward compatibility is defined as the proportion of samples that are correctly predicted by the new model among those that were correctly predicted by the old model.
By training new models to improve backward compatibility, user trust in predictions can be preserved across model updates.
Srivastava et al. empirically analyzed how naive model updates degrade backward compatibility and proposed an alternative metric that focuses on incorrect predictions rather than correct ones~\cite{BackwardCompatibility-Srivastava2020}.
In addition, generalized backward compatibility metrics have been proposed from the perspectives of true positives and false negatives~\cite{BackwardCompatibility-Sakai2022}, as well as robustness-enhanced metrics that take local data distributions into account~\cite{BackwardCompatibility-Matsuno2023}.

While the aforementioned studies define backward compatibility from the perspective of changes in prediction results, Matsuno et al. focused on SHAP-based explanations~\cite{Explainability-SHAP-Lundberg2017} and defined \emph{Backward Compatibility in eXplanation (BCX)} from the perspective of changes in explanations~\cite{BackwardCompatibility-Exp-Matsuno2024}.
This definition takes into account the concept of disagreement measures proposed in prior literature~\cite{Explainability-DisagreementProblem-Krishna2024}.
Furthermore, since Matsuno et al. applied BCX to differentiable machine learning models, an extended method has also been proposed to handle gradient boosting tree ensembles~\cite{BackwardCompatibility-Exp-Yamakura2025}.
However, BCX is currently the only existing backward compatibility metric defined from the perspective of changes in explanations.
This indicates a clear need for dedicated backward compatibility metrics suitable for explainable models that primarily rely on non-SHAP explanations, such as decision trees~\cite{CART-Breiman1984}.

Another line of related work is incremental decision trees such as Hoeffding Trees (HT)~\cite{HoeffdingTree}, also known as Very Fast Decision Trees (VFDT)~\cite{VFDT}, and Hoeffding Adaptive Trees (HAT)~\cite{HoeffdingAdaptiveTree}.
Incremental decision trees achieve practical performance on streaming environments by sequentially determining and updating branching conditions based on statistical guarantees such as the Hoeffding bound.
While there is a significant difference between update scenarios of ours and incremental decision trees in terms of whether data is processed in batch or stream format, we can expect that incremental decision trees exhibit relatively high backward compatibility with recently observed data, as they are updated sequentially in response to incoming data streams. 
However, our experiments suggest that this property does not necessarily extend to batch-wise evaluation over medium- to long-term periods.

%%%%%%%%%%%%%%%%%%%%%%%%%%%%%%%%%%%%%%%%%%%%%%%%%%
%%%%% Problem Formulation
%%%%%%%%%%%%%%%%%%%%%%%%%%%%%%%%%%%%%%%%%%%%%%%%%%
\section{Proposed Loss Metrics}
We consider updating a decision tree trained on an old dataset to a new decision tree using a new dataset, and propose \emph{Backward Compatibility Loss in Tree-based eXplanations (BCLTX)} as a loss metric.
Each sample consists of an $m$-dimensional feature vector $\bm{x} = (x_j)_{j=1}^m \in \mathbb{R}^m$ and a target variable $y$, where $y \in \{1, \ldots, c\}$ for classification tasks with $c$ classes, and $y \in \mathbb{R}$ for regression tasks.
Let the old dataset be denoted as $D_1 = \{(\bm{x}_i, y_i)\}_{i=1}^{n_1}$ and the additional dataset as $D_\Delta = \{(\bm{x}_i, y_i)\}_{i=n_1+1}^{n_2}$, and define the new dataset as $D_2 = D_1 \cup D_\Delta$.
Let $T_1$ denote the old decision tree trained on the old dataset $D_1$, and let $T_2$ denote the new decision tree updated using the new dataset $D_2$.
BCLTX is designed to capture the differences between the explanations produced by $T_1$ and $T_2$.

\subsection{Tree-Based Explanations}
A decision tree $T$ is a binary tree that, given an input $\bm{x}$, produces both a prediction of the target value and an explanation supporting the prediction.
For each node $t$ in $T$, if $t$ is an internal node, it is associated with a feature index $j_t$ and a split point $v_t$; if $t$ is a leaf node, it is associated with a predicted value $y_t$.
Given an input $\bm{x}$, inference in $T$ starts from the root node.
At each internal node $t$, the condition $x_{j_t} \leq v_t$ determines whether the inference proceeds to the left or right child, and the predicted value associated with the reached leaf node is output.
Each leaf node corresponds to a feature subspace that does not overlap with those of other leaf nodes, and the decision tree can output this subspace as an explanation for the prediction.
Algorithm~\ref{alg:predict} describes how to compute both the prediction and the explanation of a decision tree $T$ for a given input $\bm{x}$.
The procedure runs in $O(m+d)$ time, where $d$ is the depth of $T$.
Here, an explanation is represented by a lower bound vector $l = (l_j(T, \bm{x}))_{j=1}^m$ and an upper bound vector $u = (u_j(T, \bm{x}))_{j=1}^m$.

\begin{algorithm}[t]
\caption{$\textrm{PredictAndExplanation}(T,\bm{x})$}
\label{alg:predict}
\begin{algorithmic}[1]
    \STATE $l \leftarrow (-\infty)_{j=1}^m$, $u \leftarrow (\infty)_{j=1}^m$
    \STATE Let $t$ be the root node of $T$
    \WHILE{$t$ is not a leaf node}
        \IF{$x_{j_t} \leq v_t$}
            \STATE $u_{j_t} \leftarrow v_t$
            \STATE $t \leftarrow \text{The left child node of $t$}$
        \ELSE
            \STATE $l_{j_t} \leftarrow v_t$
            \STATE $t \leftarrow \text{The right child node of $t$}$
        \ENDIF
        % \IFLINE{$x_{j_t} \leq v_t$}{$u_{j_t} \leftarrow v_t$, $t \leftarrow \text{The left child node of $t$}$}
        % \ELSELINE{$l_{j_t} \leftarrow v_t$, $t \leftarrow \text{The right child node of $t$}$}
    \ENDWHILE
    \RETURN ($y_t$, $l$, $u$)
\end{algorithmic}
\end{algorithm}

We formally interpret the explanation $l, u$ produced by a decision tree $T$ for an input $\bm{x}$.
First, a feature is considered to essentially contribute to the explanation if at least one of its lower or upper bounds is finite.
The set of such features can be expressed as
\begin{align}
\label{eq:F}
    F(T,\bm{x}) &:= \{j \in \{1,\ldots,m\} \mid l_j(T,\bm{x}) > -\infty~\vee~u_j(T,\bm{x}) < \infty\}.
\end{align}
This set can be decomposed into the feature sets that contribute to the lower bounds and to the upper bounds, respectively:
\begin{align}
\begin{aligned}
\label{eq:Flu}
    F_l(T,\bm{x}) &:= \{j \in \{1,\ldots,m\} \mid l_j(T,\bm{x}) > -\infty\},\\
    F_u(T,\bm{x}) &:= \{j \in \{1,\ldots,m\} \mid u_j(T,\bm{x}) < \infty\}.\\
\end{aligned}
\end{align}
The explanation $l, u$ can be represented as the following logical expression over the feature set $F(T, \bm{x})$: a conjunction of intervals defined by the lower and upper bounds of each contributing feature.
\begin{align}
\label{eq:E}
    E(T,\bm{x}) := \bigwedge_{j \in F(T,\bm{x})} \left( l_j(T,\bm{x}) < j\text{-th feature} \leq u_j(T,\bm{x}) \right).
\end{align}

\subsection{Definitions of BCLTX}
Motivated by the tree-based explanations described above, we propose four variants of BCLTX.
Two of them are based on Eqs.~\eqref{eq:F} and~\eqref{eq:Flu}, and the remaining two are based on Eq.~\eqref{eq:E}.
These definitions are formulated as basic measures derived from commonly used dissimilarity metrics.

First, based on Eq.~\eqref{eq:F}, we define BCLTX by measuring the difference between $F(T_1, \bm{x})$ and $F(T_2, \bm{x})$ using the Jaccard coefficient.
\begin{align*}
    \delta_{\mathrm{JAC}}(T_1,T_2,\bm{x}) := \frac{1}{m}|F(T_1,\bm{x}) \bigtriangleup F(T_2,\bm{x})|
\end{align*}
Here, $\bigtriangleup$ denotes the symmetric difference operator between sets.
This definition of BCLTX reflects the extent to which the features contributing to the explanations differ.
Next, this Jaccard-based definition can be extended to a formulation using Eq.~\eqref{eq:Flu}:
\begin{align*}
\begin{aligned}
    &\delta_{\mathrm{JLU}}(T_1,T_2,\bm{x}) := \\
    &\quad \frac{1}{2m} \left( |F_l(T_1,\bm{x}) \bigtriangleup F_l(T_2,\bm{x})| + |F_u(T_1,\bm{x}) \bigtriangleup F_u(T_2,\bm{x})| \right).
\end{aligned}
\end{align*}
This captures the differences in the features contributing to the explanations with respect to the lower and upper bounds separately.

The definitions of BCLTX based on Eq.~\eqref{eq:E} quantify differences between the feature-wise intervals of $E(T_1, \bm{x})$ and $E(T_2, \bm{x})$ using the mean absolute error (MAE) or the mean squared error (MSE).
To this end, two preprocessing steps are applied.
First, to ensure finite evaluation of interval differences, we use adjusted bounds $l'_j(T,\bm{x}) := \max\{ l_j(T,\bm{x}),x_j^{\mathrm{min}}\}$ and $u'_j(T,\bm{x}) := \min\{u_j(T,\bm{x}),x_j^{\mathrm{max}}\}$, where $x_j^{\mathrm{min}} := \min\{x_{i,j}\}_{i=1}^{n_2}$ and $x_j^{\mathrm{max}} := \max\{x_{i,j}\}_{i=1}^{n_2}$.
Second, to achieve scale-invariant evaluation across features, we introduce the standard deviation $\sigma_j$ of $\{x_{i,j}\}_{i=1}^{n_2}$.
Under these preprocessing steps, BCLTX based on MAE and MSE are defined as follows.
\begin{align*}
\begin{aligned}
    &\delta_{\mathrm{MAE}}(T_1,T_2,\bm{x}) :=\\
    &\quad \frac{1}{2m} \sum_{j=1}^m \frac{|l'_j(T_1,\bm{x}) - l'_j(T_2,\bm{x})| + |u'_j(T_1,\bm{x}) - u'_j(T_2,\bm{x})|}{\sigma_j}
\end{aligned}\\
\begin{aligned}
    &\delta_{\mathrm{MSE}}(T_1,T_2,\bm{x}) :=\\
    &\quad \frac{1}{2m} \sum_{j=1}^m \frac{(l'_j(T_1,\bm{x}) - l'_j(T_2,\bm{x}))^2 + (u'_j(T_1,\bm{x}) - u'_j(T_2,\bm{x}))^2}{\sigma_j^2}
\end{aligned}
\end{align*}

%%%%%%%%%%%%%%%%%%%%%%%%%%%%%%%%%%%%%%%%%%%%%%%%%%
%%%%% Decision Tree Update under BCLTX
%%%%%%%%%%%%%%%%%%%%%%%%%%%%%%%%%%%%%%%%%%%%%%%%%%
\section{Decision Tree Update under BCLTX}
We formulate the problem of updating an old decision tree $T_1$ to a new decision tree $T_2$ on the new training dataset $D_2$ so as to suppress BCLTX.
Furthermore, we propose an efficient algorithm to address the problem, \emph{CART with Backward Compatibility in Tree-based eXplanations (CART-BCTX)}, which enhances CART~\cite{CART-Breiman1984}, a standard algorithm for constructing decision trees.

\subsection{Formulation}
Our decision tree update problem aims to minimize the following objective function using a hyperparameter $\lambda \in \mathbb{R}_{\geq 0}$ that controls the strength of BCLTX.
\begin{align}
\label{eq:objective}
\begin{aligned}
    &f(T_1,T_2,D_2,\lambda) :=\\
    &\quad \frac{1}{|D_2|} \sum_{(\bm{x},y) \in D_2} \ell(T_2(\bm{x}),y) + \frac{\lambda}{|D_s|} \sum_{(\bm{x},y) \in D_s} \delta(T_1,T_2,\bm{x})
\end{aligned}
\end{align}
Here, $T(\bm{x})$ denotes the predicted value on the decision tree $T$ for the input $\bm{x}$, $\ell$ denotes a task-specific loss function, such as the 0-1 loss for classification tasks or the squared error for regression tasks, $\delta \in \{\delta_\mathrm{JAC},\delta_\mathrm{JLU},\delta_\mathrm{MAE},\delta_\mathrm{MSE}\}$, and $D_s$ denotes the set of samples that are correctly predicted by $T_1$.
Specifically, $D_s$ is defined as
\begin{align*}
    D_s := \{(\bm{x},y) \in D_2 \mid s(T_1(\bm{x}),y) = 1\}
\end{align*}
where $s$ is an indicator function defined as
\begin{align*}
    s(\hat{y},y) := \begin{cases}
        \bm{1}[\hat{y} = y] & (\text{classification}),\\
        \bm{1}[|\hat{y} - y| \leq \epsilon] & (\text{regression}).
    \end{cases}
\end{align*}
Here, for regression tasks, a prediction is regarded as correct if the absolute error is less than or equal to a given threshold $\epsilon \in \mathbb{R}_{\geq 0}$.

The above definition incorporates backward compatibility into the loss function by weighting only samples for which the old model made correct predictions.
In practice, however, the definition of $D_s$ is flexible and can be adjusted to meet user requirements.
For example, $D_s$ can be defined as the set of samples that satisfy a decision rule in which the user has sufficient confidence, enabling model updates that better reflect the user’s intentions.
Furthermore, it is also possible to assign more fine-grained real-valued weights to individual samples.

We need to develop an efficient algorithm to address the above problem.
However, this problem is computationally challenging even when $\lambda = 0$~\cite{Hardness-Hyafil1976}.
Therefore, we enhance CART~\cite{CART-Breiman1984}, a greedy algorithm for constructing decision trees, so that it can handle BCLTX.
In addition, although decision trees employ a post-pruning method known as cost complexity pruning~\cite{CART-Breiman1984}, this pruning method can be readily extended to handle BCLTX.

\subsection{CART-BCTX}
In this subsection, we describe CART-BCTX in detail.
CART-BCTX is a greedy decision tree update algorithm that enhances CART to heuristically optimize Eq.~\eqref{eq:objective}.
CART starts from a minimal decision tree consisting of only a root node and grows the tree by recursively selecting greedy splits that improve the objective, as long as given constraints on the maximum depth and the minimum sample size are satisfied.
CART-BCTX incorporates operations that account for BCLTX into the greedy split selection procedure of CART.
Specifically, CART-BCTX maintains, for each node in the growing decision tree, a lower bound vector $\hat{l}$ and an upper bound vector $\hat{u}$ that represent the feature subspace corresponding to that node.
The most greedy split is then searched using $\hat{l}$ and $\hat{u}$.
Pseudo-code corresponding to the recursive procedure for growing a decision tree in CART-BCTX is shown in Algorithm~\ref{alg:cart-bctx} (initially call $\textrm{CART-BCTX}(D_2,(x_j^{\mathrm{min}})_{j=1}^m,(x_j^{\mathrm{max}})_{j=1}^m,0)$ where $T_1$, $D_s$, and $\lambda$ are globally stored).
In the following, we describe in detail the procedure searching greedy splits (Algorithm~\ref{alg:split}) and present the resulting computational complexity of CART-BCTX.

\begin{algorithm}[t]
\caption{$\textrm{CART-BCTX}(D,\hat{l},\hat{u},d)$}
\label{alg:cart-bctx}
\begin{algorithmic}[1]
    \STATE Generate a node $t$
    \STATE $y_t \leftarrow \text{Generate a prediction result on $D$}$
    \IF{$d = \text{max\_depth}$}
        \RETURN The tree with only the root node $t$
    \ENDIF
    % \IFRET{$d = \text{max\_depth}$}{The tree with only the node $t$}
    \STATE $(j_t, v_t) \leftarrow \textrm{FindBestSplit-BCTX}(D,\hat{l},\hat{u})$
    \IF{$j_t = \mathrm{NULL}$}
        \RETURN The tree with only the root node $t$
    \ENDIF
    % \IFRET{$j_t = \mathrm{NULL}$}{The tree with only the node $t$}
    \STATE $D_L \leftarrow \{(\bm{x},y) \in D \mid x_{j_t} \leq v_t\}$
    \STATE $D_R \leftarrow \{(\bm{x},y) \in D \mid x_{j_t} > v_t\}$
    \STATE $\check{l} \leftarrow \hat{l}$, $\check{u} \leftarrow \hat{u}$
    \STATE $\check{l}_{j_t} \leftarrow v_{j_t}$, $\check{u}_{j_t} \leftarrow v_{j_t}$,
    \STATE $T_L \leftarrow \textrm{CART-BCTX}(D_L,\hat{l},\check{u},d+1)$
    \STATE $T_R \leftarrow \textrm{CART-BCTX}(D_R,\check{l},\hat{u},d+1)$
    \RETURN The tree rooted at $t$ with the left subtree $T_L$ and the right subtree $T_R$
\end{algorithmic}
\end{algorithm}

\begin{algorithm}[t]
\caption{$\textrm{FindBestSplit-BCTX}(D,\hat{l},\hat{u})$}
\label{alg:split}
\begin{algorithmic}[1]
    \STATE $j^* \leftarrow \mathrm{NULL}$, $v^* \leftarrow \mathrm{NULL}$, $\mathrm{Gain}^* \leftarrow 0$
    \FOR{$j = 1, \ldots, m$}
        \STATE Compute threshold candidates $V$ of $j$-th feature over $D$
        \STATE $D_L \leftarrow \emptyset$, $D_R \leftarrow D$
        \FOR{$v \in V$ in ascending order}
            \STATE $D_\mathrm{diff} \leftarrow \{(\bm{x},y) \in D_R \mid x_j \leq v\}$
            \STATE $D_L \leftarrow D_L \cup D_\mathrm{diff}$, $D_R \leftarrow D_R \setminus D_\mathrm{diff}$
            \STATE Compute $\mathrm{Gain}_\ell$ and $\mathrm{Gain}_\delta$ of $(D_L,D_R)$
            \IFLINE{$|D_L| < \text{min\_samples\_leaf}$}{\textbf{continue}}
            \IFLINE{$|D_R| < \text{min\_samples\_leaf}$}{\textbf{continue}}
            \IF{$\mathrm{Gain}^* < \frac{1}{|D_2|}\mathrm{Gain}_\ell + \frac{\lambda}{|D_s|} \mathrm{Gain}_\delta$}
                \STATE $j^* \leftarrow j$, $v^* \leftarrow v$, $\mathrm{Gain}^* \leftarrow \frac{1}{|D_2|}\mathrm{Gain}_\ell + \frac{\lambda}{|D_s|} \mathrm{Gain}_\delta$
            \ENDIF
        \ENDFOR
    \ENDFOR
    \RETURN $(j^*,v^*)$
\end{algorithmic}
\end{algorithm}

Let $D$ denote the dataset associated with the node currently being considered for splitting.
For each feature $j$, the set of candidate split points $V$ is defined as the midpoints between consecutive values when the samples in $D$ are sorted in ascending order with respect to $x_j$.
For each split point $v \in V$, which is scanned in ascending order, we compute the gain with respect to the task-specific loss, denoted by $\mathrm{Gain}_\ell$, and the gain with respect to BCLTX, denoted by $\mathrm{Gain}_\delta$.
Among all combinations of features and split points, we select the split that maximizes the total gain $\frac{1}{|D_2|}\mathrm{Gain}_\ell + \frac{\lambda}{|D_s|} \mathrm{Gain}_\delta$.
If the total gain does not exceed zero for any split, no split is performed.
Since $\mathrm{Gain}_\ell$ can be computed in the same manner as in CART, we omit its description and focus only on how to efficiently compute $\mathrm{Gain}_\delta$.
The computation of $\mathrm{Gain}_\delta$ depends on the variant of the BCLTX definition being used.
Here, when the dataset $D$ is split by a point $v$ on feature $j$, we denote by $D_L$ and $D_R$ the sets of samples in $D$ satisfying $x_j \leq v$ and $x_j > v$, respectively.
In addition, we denote by $D_\mathrm{diff}$ the set of samples that lie between the split point $v$ and the immediately preceding split point.
A key observation is that the effect of a split using $j$ and $v$ on the explanations is solely that, for samples in $D_L$, the upper bound $\hat{u}_j$ changes to $v$, whereas for samples in $D_R$, the lower bound $\hat{l}_j$ changes to $v$.

\paragraph{For $\delta_\mathrm{JAC}$}
We need to check whether the feature $j$ currently under consideration has already been used.
Let $\bar{l}_j := \bm{1}[\hat{l}_j = x_j^{\mathrm{min}}]$ and $\bar{u}_j := \bm{1}[\hat{u}_j = x_j^{\mathrm{max}}]$ for convenience to represent $j$ is not used for $\hat{l}$ and $\hat{u}$, respectively.
We can compute the gain
\begin{align*}
    \mathrm{Gain}_{\delta_\mathrm{JAC}} = \frac{1}{m}\sum_{(\bm{x},y) \in D \cap D_s} \bar{l}_j \bar{u}_j (\bm{1}[j \in F(T_1,\bm{x})] - \bm{1}[j \notin F(T_1,\bm{x})]).
\end{align*}
This gain depends only on the feature $j$ and is therefore constant with respect to the choice of split point $v$.

\paragraph{For $\delta_\mathrm{JLU}$}
Similarly to $\delta_\mathrm{JAC}$, we need to check whether the feature $j$ currently under consideration has already been used for the upper and lower bounds, respectively.
We can compute the gain
\begin{align*}
\begin{aligned}
    \mathrm{Gain}_{\delta_\mathrm{JLU}} &= \frac{1}{2m} \sum_{(\bm{x},y) \in D_L \cap D_s} \bar{u}_j (\bm{1}[j \in F_u(T_1,\bm{x})] - \bm{1}[j \notin F_u(T_1,\bm{x})])\\
    &+ \frac{1}{2m} \sum_{(\bm{x},y) \in D_R \cap D_s} \bar{l}_j (\bm{1}[j \in F_l(T_1,\bm{x})] - \bm{1}[j \notin F_l(T_1,\bm{x})]).
\end{aligned}
\end{align*}
In addition, the gain change with respect to $D_\mathrm{diff}$ can be computed in $O(|D_\mathrm{diff}|)$ time by adding and subtracting constants $\bm{1}[j \in F_u(T_1,\bm{x})] - \bm{1}[j \notin F_u(T_1,\bm{x})]$ and $\bm{1}[j \in F_l(T_1,\bm{x})] - \bm{1}[j \notin F_l(T_1,\bm{x})]$ for all $(\bm{x},y) \in D_\mathrm{diff} \cap D_s$.

\paragraph{For $\delta_\mathrm{MAE}$}
According to the definition, we can compute the gain
\begin{align*}
\begin{aligned}
    \mathrm{Gain}_{\delta_\mathrm{MAE}} &= \frac{1}{2m\sigma_j} \sum_{(\bm{x},y) \in D_L \cap D_s} (|u'_j(T_1,\bm{x}) - \hat{u}_j| - |u'_j(T_1,\bm{x}) - v|)\\
    &+ \frac{1}{2m\sigma_j} \sum_{(\bm{x},y) \in D_R \cap D_s} (|l'_j(T_1,\bm{x}) - \hat{l}_j| - |l'_j(T_1,\bm{x}) - v|).
\end{aligned}
\end{align*}
For efficient gain computation, we maintain the set $\mathcal{U}$ of $u'_j$ for samples in $D_L \cap D_s$ and the set $\mathcal{L}$ of $l'_j$ for samples in $D_R \cap D_s$.
Then, it suffices to compute $\sum_{u'_j \in \mathcal{U}} |u'_j - w|$ and $\sum_{l'_j \in \mathcal{L}} |l'_j - w|$ for a given $w$.
By partitioning $\mathcal{U}$ into two subsets $\mathcal{U}_{\leq}$ and $\mathcal{U}_{>}$ according to the conditions $u'_j \leq w$ and $u'_j > w$, respectively, we can rewrite $\sum_{u'_j \in \mathcal{U}} |u'_j - w| = w|\mathcal{U}_{\leq}| - \sum_{u'_j \in \mathcal{U}_{\leq}} u'_j + \sum_{u'_j \in \mathcal{U}_{>}} u'_j - w|\mathcal{U}_{>}|$.
The same transformation applies to $\mathcal{L}$.
Such computations are achieved by maintaining $\mathcal{U}$ and $\mathcal{L}$ using self-balancing binary search trees, enabling the gain to be computed in $O(|D_\mathrm{diff}| \log |D|)$ time.

\paragraph{For $\delta_\mathrm{MSE}$}
According to the definition, we can compute the gain
\begin{align*}
\begin{aligned}
    \mathrm{Gain}_{\delta_\mathrm{MSE}} &= \frac{1}{2m\sigma_j^2} \sum_{(\bm{x},y) \in D_L \cap D_s} ((u'_j(T_1,\bm{x}) - \hat{u}_j)^2 - (u'_j(T_1,\bm{x}) - v)^2)\\
    &+ \frac{1}{2m\sigma_j^2} \sum_{(\bm{x},y) \in D_R \cap D_s} ((l'_j(T_1,\bm{x}) - \hat{l}_j)^2 - (l'_j(T_1,\bm{x}) - v)^2).
\end{aligned}
\end{align*}
Here, for a given set of values $\mathcal{W}$ and a value $w$, we can write $\sum_{w' \in \mathcal{W}} (w' - w)^2 = \sum_{w' \in \mathcal{W}} w'^2 + 2w \sum_{w' \in \mathcal{W}} w' + |\mathcal{W}| w^2$.
Therefore, the computation of $\mathrm{Gain}_{\delta_\mathrm{MSE}}$ with respect to $D_L$ can be completed in $O(|D_\mathrm{diff}|)$ time by updating $\sum_{u'_j \in \mathcal{U}} {u'}_j^2$ and $\sum_{u'_j \in \mathcal{U}} u'_j$ for the set $\mathcal{U}$ according to $D_{\mathrm{diff}} \cap D_s$.
The same procedure applies to $D_R$ and the corresponding set $\mathcal{L}$.

In general, CART preprocesses the dataset by sorting it in ascending order with respect to each feature, and this sorted order can be maintained even after the dataset is recursively partitioned.
Moreover, the computation of $\mathrm{Gain}_\ell$ can be performed in $O(|D_\mathrm{diff}|)$ time following the standard CART procedure.
Note that, for classification tasks, it is common practice to use the Gini impurity as a surrogate rather than directly using the 0–1 loss.
Considering these facts together with the discussion on the computation of $\mathrm{Gain}_\delta$, and noting that for each feature $j$ the total sum of $|D_{\mathrm{diff}}|$ over all split points is $|D|$, the following proposition holds regarding the computational complexity of CART-BCTX.

\begin{proposition}
Let $n = |D_2|$, $d_1$ be the depth of $T_1$, and $d_2$ be the depth of the updated tree.
We preprocess to sort the dataset and compute the explanation $l, u$ for each sample in $O(nm \log n + n(m+d_1))$ time.
Then, $\textrm{CART-BCTX}$ takes $O(nmd_2)$ time for $\delta \in \{\delta_{\mathrm{JAC}},\delta_{\mathrm{JLU}},\delta_{\mathrm{MSE}}\}$, same as CART, and $O(nmd_2 \log n)$ time for $\delta_{\mathrm{MAE}}$.
\end{proposition}

\subsection{Cost Complexity Pruning under BCLTX}
Cost complexity pruning (CCP) is a post-pruning method that reduces the size of a decision tree in order to prevent overfitting~\cite{CART-Breiman1984}.
However, when CCP is naively applied to a decision tree constructed by CART-BCTX, aggressive pruning may severely degrade backward compatibility in explanations.
Therefore, we introduce a simple extension of CCP that enables it to be executed under BCLTX.

Our CCP removes subtrees of $T_2$ deemed unnecessary by minimizing the following objective function.
\begin{align*}
    g(T_1,T_2^\mathrm{ccp},D_2,\lambda,\alpha) := f(T_1,T_2^\mathrm{ccp},D_2,\lambda)  + \alpha \#\mathrm{Leaves}(T_2^\mathrm{ccp})
\end{align*}
Here, $T_2^\mathrm{ccp}$ is the pruned decision tree, $\#\mathrm{Leaves}(T_2^\mathrm{ccp})$ denotes the number of its leaves, and $\alpha \in \mathrm{R}_{\geq 0}$ is a regularization parameter.
Our CCP differs from the standard CCP only in replacing the loss function $f$ with our formulation of Eq.~\eqref{eq:objective}, and introduces no changes from an algorithmic perspective.
That is, as in the standard CCP~\cite{CART-Breiman1984}, it is possible to efficiently search for candidate values of $\alpha$, compute $T_2^\mathrm{ccp}$ for each $\alpha$, and select a better $\alpha$ via cross-validation on $D_2$.

%%%%%%%%%%%%%%%%%%%%%%%%%%%%%%%%%%%%%%%%%%%%%%%%%%
%%%%% Experiments
%%%%%%%%%%%%%%%%%%%%%%%%%%%%%%%%%%%%%%%%%%%%%%%%%%
\section{Experiments}
To evaluate the performance of CART-BCTX, we investigated its behavior on real-world datasets by comparing it with a naive model update using CART as well as incremental decision trees.
CART-BCTX was implemented using C++20 and modularized in Python 3.13 via Cython 3.2.
Note that CART can be executed within CART-BCTX without any additional overhead by explicitly skipping the computation of $\mathrm{Gain}_\delta$ when $\lambda = 0$.
In addition, we used the River library~\cite{montiel2021river} for incremental decision trees.
The experimental environment consisted of Ubuntu 24.04 LTS on WSL2, 13th Gen Intel(R) Core(TM) i7-13700H CPU, and 16GB of memory.

Our experiments are designed to address the following four research questions:
\begin{enumerate}
  \item[(i)] Can CART-BCTX achieve a well-balanced trade-off between predictive performance and BCLTX values?
  \item[(ii)] Does improving BCLTX values affect backward compatibility in predictions?
  \item[(iii)] Can CART-BCTX operate with computation times comparable to those of CART?
  \item[(iv)] How does the behavior of CART-BCTX differ from that of incremental decision trees?
\end{enumerate}

\paragraph{Datasets}
We used a total of 10 datasets obtained from the UCI Machine Learning Repository~\cite{uci} and the KEEL Dataset Repository~\cite{KEEL-Alcal-Fdez2011}.
The datasets consist of five classification tasks and five regression tasks; the dataset names, numbers of samples, and numbers of features are summarized in Table~\ref{tab:datasets}.
Categorical features were preprocessed using one-hot encoding, and the reported numbers of features correspond to those after preprocessing.

\paragraph{Settings}
For both CART and CART-BCTX, no depth limit was imposed, the minimum sample size was set to 1\% of the number of samples in the training dataset, and the post-pruning process by CCP was performed using 5-fold cross-validation.
For CART-BCTX, the parameter $\lambda$ was varied over nine values: $0.0001$, $0.0005$, $0.001$, $0.005$, $0.01$, $0.05$, $0.1$, $0.5$, and $1.0$.
All variants of BCLTX were tested, namely $\delta_{\mathrm{JAC}}$, $\delta_{\mathrm{JLU}}$, $\delta_{\mathrm{MAE}}$, and $\delta_{\mathrm{MSE}}$.
We set the number of samples in the old training dataset $D_1$ to 500 and that in the new training dataset $D_2$ to 1,000, using the remaining samples as the test dataset.
The old decision tree was trained by CART on $D_1$.
For each dataset, a total of 100 trials were conducted using random splits of the dataset.
In regression tasks, the parameter $\epsilon$ used to define correct predictions by the old model was set to the root mean squared error (RMSE) of the old decision tree evaluated on $D_2$.

\begin{table}[t]
    \centering
    \begin{tabular}{ccccc}
       Task & Dataset &  \#Samples & \#Features & Source\\ \hline
       & adult & 45222 & 104 & UCI \\
       & magic & 19020 & 10 & UCI \\
       Classification & shoppers & 12330 & 28 & UCI \\
       & spambase & 4601 & 57 & UCI \\
       & thyroid & 7200 & 21 & KEEL \\ \hline
       & abalone & 4177 & 10 & UCI \\
       & ailerons & 13750 & 40 & KEEL \\
       Regression & power-plant & 9568 & 4 & UCI \\
       & puma32h & 8192 & 32 & KEEL \\
       & wine-quality & 6497 & 13 & UCI \\ \hline
    \end{tabular}
    \caption{Summary of datasets used in our experiments.}
    \label{tab:datasets}
\end{table}

\begin{figure*}[t]
    \centering
    \includegraphics[width=\textwidth]{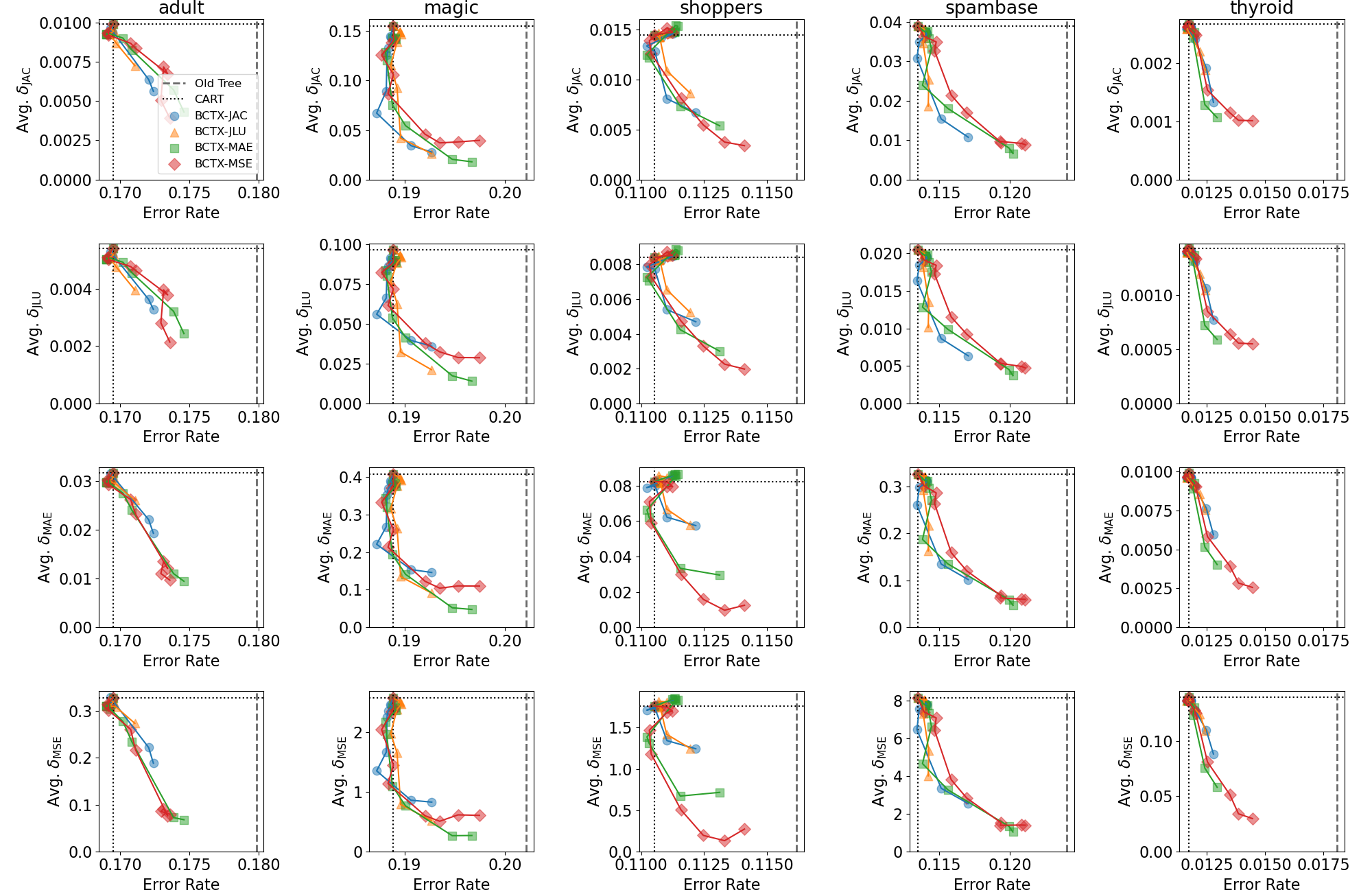}
    \caption{
        Trade-offs for classification tasks.
        BCTX-\emph{tmp} denotes CART-BCTX using $\delta_{\mathrm{tmp}}$.
        The horizontal and vertical axes represent the error rate and the sample-averaged BCLTX, respectively (lower is better for both).
        The gray dashed lines indicate the error rates of the old models.
        The gray dotted lines represent the error rates and BCLTX values achieved by CART.
        All plotted values are averaged over 100 trials.
        Overall, CART-BCTX achieves a reasonable trade-off between error rate and BCLTX.
    }
    \Description{Results for classification tasks.}
    \label{fig:classif}
\end{figure*}

\begin{figure*}[t]
    \centering
    \includegraphics[width=\textwidth]{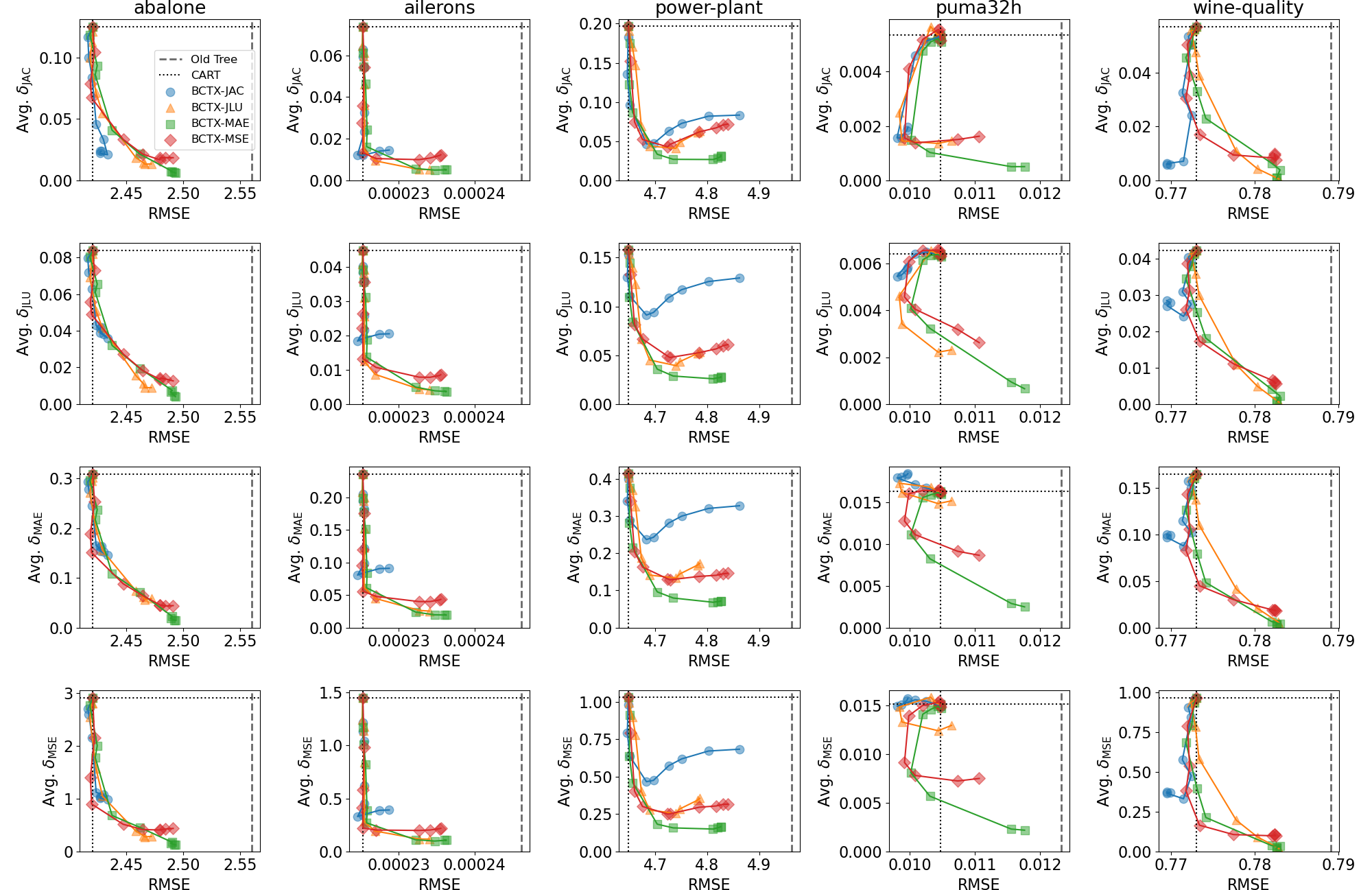}
    \caption{
        Trade-offs for regression tasks.
        BCTX-\emph{tmp} denotes CART-BCTX using $\delta_{\mathrm{tmp}}$.
        The horizontal and vertical axes represent the RMSE and the sample-averaged BCLTX, respectively (lower is better for both).
        The gray dashed lines indicate the RMSEs of the old models.
        The gray dotted lines represent the RMSEs and BCLTX values achieved by CART.
        All plotted values are averaged over 100 trials.
        Overall, CART-BCTX achieves a reasonable trade-off between RMSE and BCLTX.
    }
    \Description{Results for regression tasks.}
    \label{fig:regress}
\end{figure*}

\subsection{Prediction–BCLTX Trade-off Analysis}
Figure~\ref{fig:classif} (for classification tasks) and Figure~\ref{fig:regress} (for regression tasks) present the predictive performance and BCLTX values of the updated decision trees obtained by CART and CART-BCTX.
In these figures, the values obtained by CART serve as the starting point, and the trajectories illustrate changes in the evaluation metrics as $\lambda$ increases in CART-BCTX, where each point represents the average over 100 trials.
As reference lines, the predictive performance of the old decision tree and that of the new decision tree obtained by CART, as well as the BCLTX values obtained by CART, are shown.
As measures of predictive performance, we used the error rate (i.e., misclassification rate) for classification datasets and the RMSE for regression datasets.
For all datasets, we evaluated the sample-averaged values of $\delta_{\mathrm{JAC}}$, $\delta_{\mathrm{JLU}}$, $\delta_{\mathrm{MAE}}$, and $\delta_{\mathrm{MSE}}$ over samples that were correctly predicted by the old decision tree.

From both Figure~\ref{fig:classif} and Figure~\ref{fig:regress}, we observe that CART-BCTX achieves improvements in predictive performance comparable to those of CART for small $\lambda$, and that the magnitude of performance improvement decreases as $\lambda$ increases.
In addition, compared with CART, CART-BCTX consistently improves BCLTX values, while the improvement in BCLTX becomes more gradual as $\lambda$ increases.
These results indicate that CART-BCTX exhibits a reasonable trade-off between improvements in predictive performance and improvements in BCLTX as $\lambda$ varies.
Although the behavior of CART-BCTX varies depending on the variant of BCLTX used during execution, employing $\delta_{\mathrm{MAE}}$ and $\delta_{\mathrm{MSE}}$ leads to notably improved values of the other loss metrics, $\delta_{\mathrm{JAC}}$ and $\delta_{\mathrm{JLU}}$, across all datasets.
That is, using $\delta_{\mathrm{MAE}}$ or $\delta_{\mathrm{MSE}}$ is sufficient when executing CART-BCTX.
In summary, \emph{CART-BCTX is capable of balancing the trade-off between predictive performance and BCLTX values, and execution with $\delta_{\mathrm{MAE}}$ or $\delta_{\mathrm{MSE}}$ is recommended in most cases}.

\subsection{Effect on Compatible Predictions}
\begin{figure*}[t]
    \centering
    \includegraphics[width=\textwidth]{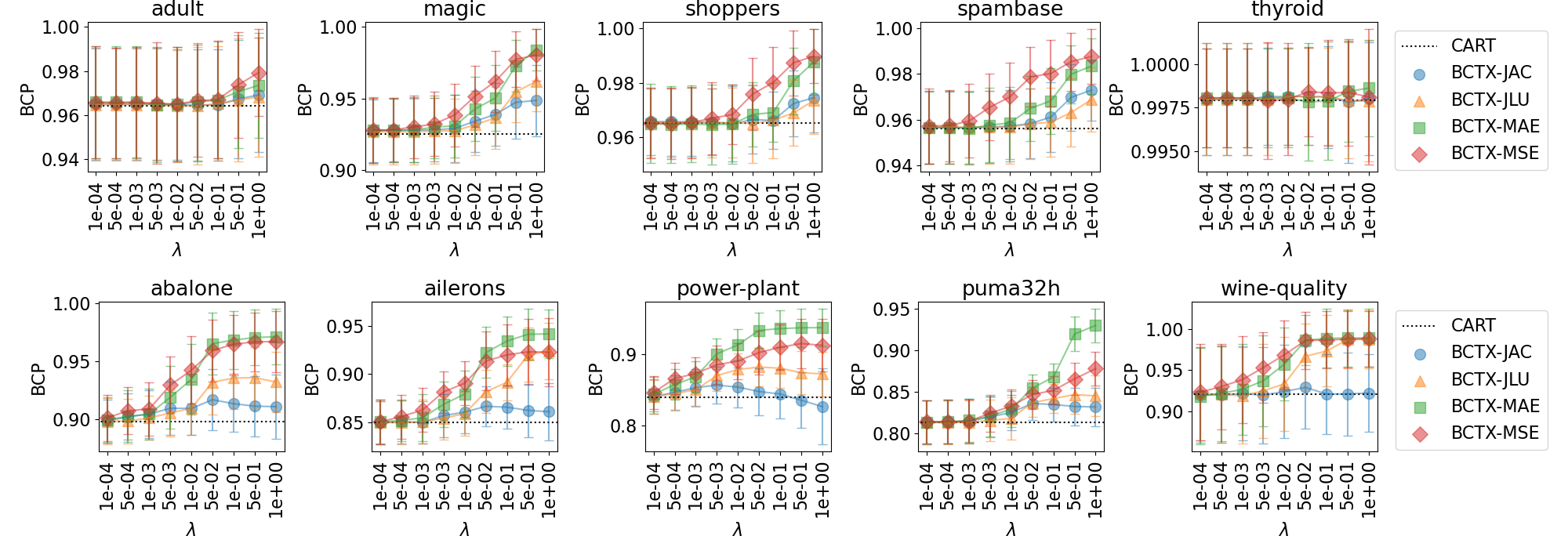}
    \caption{
        Resulting BCP values (higher is better).
        BCTX-\emph{tmp} denotes CART-BCTX using $\delta_{\mathrm{tmp}}$.
        The plotted values represent the mean over 100 trials with the standard deviation.
        CART-BCTX with $\delta_{\mathrm{MAE}}$ or $\delta_{\mathrm{MSE}}$ achieves higher BCP as $\lambda$ increases.
    }
    \Description{Results for BCP.}
    \label{fig:bcpr}
\end{figure*}

Although CART-BCTX focuses on backward compatibility in explanations, it is also desirable in practice that it maintains Backward Compatibility in Predictions (BCP)~\cite{BackwardCompatibility-Bansal2019}.
Figure~\ref{fig:bcpr} shows the BCP baseline obtained by CART and the changes in BCP achieved by CART-BCTX with varying $\lambda$.
Here, BCP is defined as the proportion of samples that are correctly predicted by the new decision tree among those that are correctly predicted by the old decision tree:
\begin{align*}
    \mathrm{BCP}(T_1,T_2,D_\mathrm{test}) := \frac{\sum_{(\bm{x},y) \in D_\mathrm{test}} s(T_1(\bm{x}),y) s(T_2(\bm{x}),y)}{\sum_{(\bm{x},y) \in D_\mathrm{test}} s(T_1(\bm{x}),y)}.
\end{align*}

From Figure~\ref{fig:bcpr}, we observe that when CART-BCTX is executed using $\delta_{\mathrm{MAE}}$ or $\delta_{\mathrm{MSE}}$, BCP improves substantially as $\lambda$ increases.
In contrast, when $\delta_{\mathrm{JAC}}$ or $\delta_{\mathrm{JLU}}$ is used, cases in which BCP improves are relatively limited.
Note that for the `thyroid` dataset, the BCP achieved by CART is already sufficiently high, and therefore little improvement is obtained by CART-BCTX.
Intuitively, this behavior can be attributed to the fact that improving $\delta_{\mathrm{MAE}}$ or $\delta_{\mathrm{MSE}}$ reduces changes in how samples are partitioned by the decision tree, making the predictions of the new decision tree more likely to be equivalent to those of the old decision tree.
In summary, \emph{CART-BCTX is also effective in improving backward compatibility in predictions, and execution with $\delta_{\mathrm{MAE}}$ or $\delta_{\mathrm{MSE}}$ is recommended in most cases}.

\subsection{Computation Times}
Theoretically, the computational complexity of CART-BCTX is comparable to that of CART.
However, in practice, additional computations related to BCLTX introduce a certain amount of overhead.
Therefore, to evaluate the extent of this overhead, we compared the computation times of CART and CART-BCTX.

\begin{figure*}[t]
    \centering
    \includegraphics[width=\textwidth]{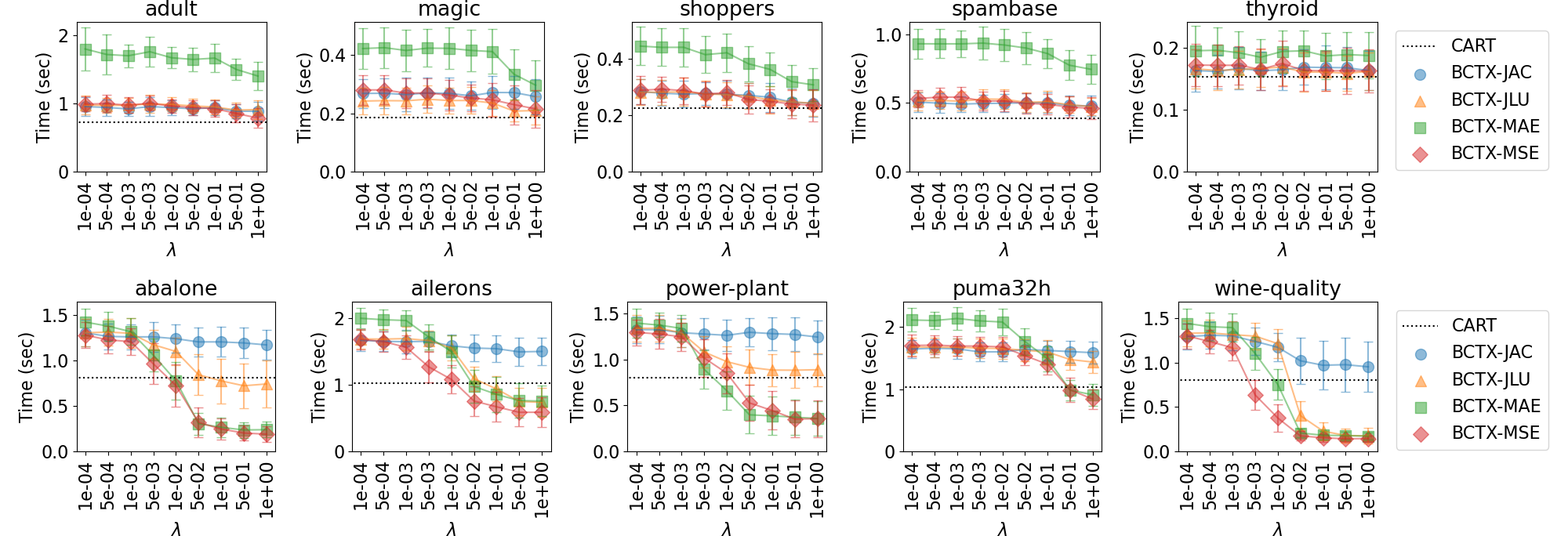}
    \caption{
        Resulting computation times.
        BCTX-\emph{tmp} denotes CART-BCTX using $\delta_{\mathrm{tmp}}$.
        The plotted values represent the mean over 100 trials with the standard deviation.
        The computation time of CART-BCTX is comparable to that of CART.
    }
    \Description{Results for computation times.}
    \label{fig:time}
\end{figure*}

Figure~\ref{fig:time} shows the results.
CART-BCTX using $\delta_{\mathrm{JAC}}$, $\delta_{\mathrm{JLU}}$, or $\delta_{\mathrm{MSE}}$ achieves computation times comparable to those of CART, and at worst less than twice.
In addition, although CART-BCTX using MAE theoretically incurs a worst-case overhead of $\log n$, the empirical results show that, for $n = 1000$, the computation times are at most approximately twice those of CART.
For regression tasks, cases were observed in which CART-BCTX is faster than CART for large $\lambda$.
This can be attributed to the fact that the computational complexity depends on the depth of the constructed decision tree, and that incorporating BCLTX naturally suppresses unnecessary splits.
In summary, \emph{CART-BCTX operates with computation times that are practically comparable to those of CART}.

\begin{figure*}[t]
    \centering
    \subfigure{\includegraphics[width=0.49\textwidth]{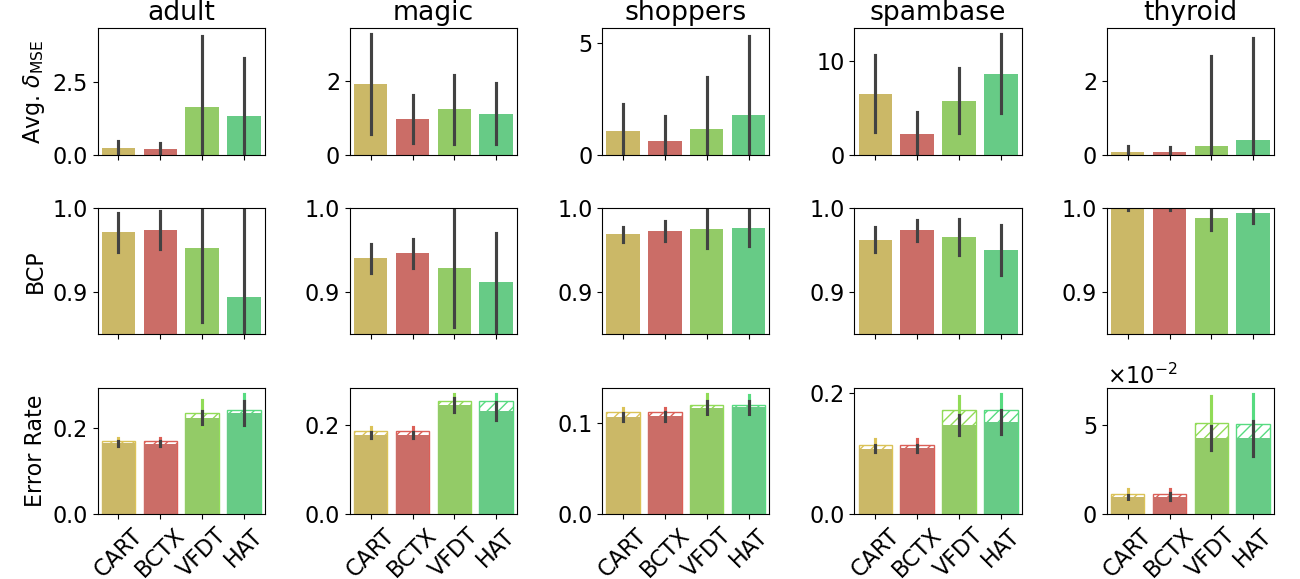}}
    \subfigure{\includegraphics[width=0.49\textwidth]{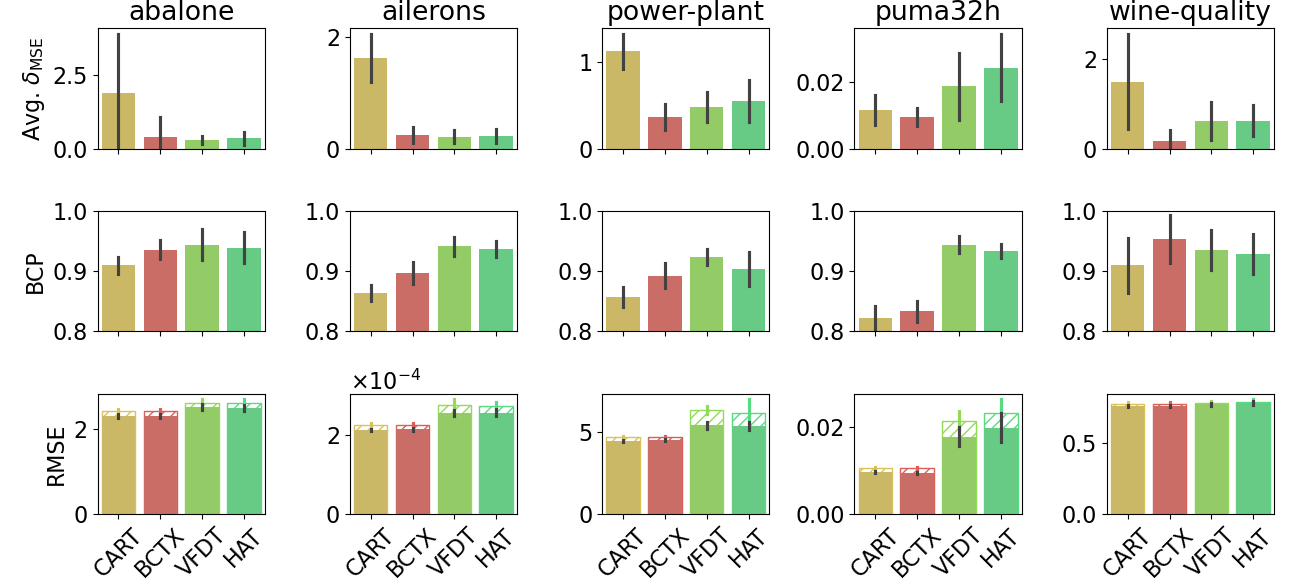}}
    \caption{
        sample-averaged BCLTX ($\delta_\mathrm{MSE}$), BCP, and loss values for CART, CART-BCTX ($\lambda=0.005$, $\delta=\delta_\mathrm{MSE}$), VFDT, and HAT.
        The plotted values represent the mean over 100 trials with the standard deviation.
        The hatched areas mean the improvement degrees of the losses.
        CART-BCTX offered a better balance of backward compatibility and accuracy than others in many cases.
    }
    \Description{Results for vs. incremental decision trees.}
    \label{fig:vs-stream-learner}
\end{figure*}

\subsection{Comparison with Incremental Trees}
\label{sec:comparison-with-incremental-trees}
We investigated the differences in behavior between CART-BCTX and the incremental decision trees VFDT~\cite{VFDT} and HAT~\cite{HoeffdingAdaptiveTree} using the same dataset as in the above experiments.
Since VFDT and HAT are designed to assume a sufficient number of samples, we changed the sample sizes to $|D_1| = 1000$ and $|D_2| = 2000$.
Furthermore, we used the implementations in the River library~\cite{montiel2021river}~\footnote{\url{https://riverml.xyz/latest/}} for VFDT and HAT, setting the parameters to ``grace\_period=50'' and ``tau=0.15'', and for HAT, ``drift\_window\_threshold = 100''.
For CART-BCTX, we used the settings $\lambda = 0.005$ and $\delta = \delta_\mathrm{MSE}$, which achieved a good prediction-BCLTX trade-off in many cases in the previous experiments.
Under these settings, we measured the prediction loss and the BCP for CART, CART-BCTX, VFDT, and HAT, as well as the sample-averaged BCLTX ($\delta_\mathrm{MSE}$) for samples correctly predicted by the old decision trees.
In addition, Appendix~\ref{sec:appendix-visualized-tree-comparizon} provides specific examples of old and new decision trees obtained using each method on the spambase dataset.

Note that incremental decision trees are not explicitly designed to enforce backward compatibility in both predictions and explanations.
Our comparison aims to highlight differences in their behavior under batch-wise evaluation of backward compatibility, while also comparing prediction performance.

Figure~\ref{fig:vs-stream-learner} shows the results.
Regarding the BCLTX values based on $\delta_\mathrm{MSE}$, CART-BCTX clearly achieved the best results in many datasets except for abalone and ailerons.
For abalone and ailerons datasets, CART-BCTX, VFDT, and HAT all achieved comparable results.
Regarding BCP values, CART-BCTX consistently achieved good results in all cases for the classification tasks, whereas VFDT and HAT consistently showed good results for the regression tasks.
However, with the exception of the puma32h dataset, CART-BCTX also achieved relatively good BCP values on the regression tasks.
Based on these findings, when observing performance on a batch-wise basis, incremental decision trees maintained a certain level of backward compatibility in predictions, but were inferior in terms of backward compatibility in tree-based explanations compared to CART-BCTX.
Regarding prediction loss, the CART variants clearly exhibited lower values than incremental decision trees, reflecting the superior fit of batch learning on non-large datasets.
In summary, \emph{CART-BCTX is superior to incremental decision trees in both of predictive performance and BCLTX values, and has a comparable performance in BCP values, when measuring batch-wise performances}.

%%%%%%%%%%%%%%%%%%%%%%%%%%%%%%%%%%%%%%%%%%%%%%%%%%
%%%%% Conclusion
%%%%%%%%%%%%%%%%%%%%%%%%%%%%%%%%%%%%%%%%%%%%%%%%%%
\section{Conclusion}
In this paper, we proposed Backward Compatibility Loss in Tree-based eXplanations (BCLTX), a loss metric that captures backward compatibility with respect to explanations in decision tree updates.
We also proposed CART with Backward Compatibility in Tree-based eXplanations (CART-BCTX), an efficient greedy algorithm for the decision tree update problem under BCLTX.
CART-BCTX has computational complexity comparable to that of the standard decision tree construction algorithm CART.
Experimental results on 10 real-world datasets covering both classification and regression tasks demonstrated that CART-BCTX achieves favorable trade-offs between predictive performance and BCLTX, with computation times comparable to those of CART.
In addition, we confirmed that strengthening backward compatibility in explanations through BCLTX also leads to improved backward compatibility in predictions.
Furthermore, we confirmed that CART-BCTX achieves superior performance in both predictive accuracy and BCLTX compared with incremental decision trees under batch-wise evaluation.
We believe that our work supports the trustworthy deployment of decision trees in risk-sensitive decision-making tasks, where trust in model predictions is critically important.

\paragraph{Limitations and Future Work}
There are three main future directions.
First, since constructing an optimal decision tree is computationally intractable in general~\cite{Hardness-Hyafil1976} (i.e., even in $\lambda = 0$ cases for our situation), CART-BCTX is designed as a greedy algorithm.
From a practical perspective, it would be important to extend recent efficient algorithms for learning near-optimal decision trees~\cite{ODT-Babbar2025,ODT-Kohler2025} so that they can handle BCLTX.
Second, BCLTX does not explicitly consider model updates under concept drift~\cite{Drift-Lu2019}.
Under concept drift, it is necessary to discuss for which samples backward compatibility in explanations can or should be preserved.
We report preliminary experimental results of CART-BCTX under concept drift scenarios in Appendix~\ref{sec:under-concept-drift}.
Finally, it remains to be investigated to what extent strengthening backward compatibility in explanations through BCLTX is meaningful in practice.
To this end, conducting user studies in realistic deployment settings, including synthetic environments such as those considered by Bansal et al.~\cite{BackwardCompatibility-Bansal2019}, would be an important direction for future work.

\begin{acks}
This work was supported by Japan Science and Technology Agency (JST), ACT-X Grant Number JPMJAX24CE.
\end{acks}

\bibliographystyle{ACM-Reference-Format}
\balance
\bibliography{kdd26}

\appendix

\section{Details of Our Cost Complexity Prunig}
Here, we describe CCP under BCLTX in detail.
Our CCP minimizes the following objective function.
\begin{align}
\label{eq:ccp}
    g(T_1,T_2^\mathrm{ccp},D_2,\lambda,\alpha) := f(T_1,T_2^\mathrm{ccp},D_2,\lambda)  + \alpha \#\mathrm{Leaves}(T_2^\mathrm{ccp})
\end{align}
In CCP, for each node $t$ in $T_2$, we can compute the cost $\alpha_t$ required to remove the subtree rooted at $t$ and make $t$ a new leaf node.
Let the set of samples associated with node $t$ be denoted as
\begin{align*}
    D_t := \{(\bm{x},y) \in D_2 \mid \bm{x}~\text{reaches the node}~t\}
\end{align*}
Then, the value of the loss function $f$ restricted to the subtree rooted at $t$ can be computed as follows.
\begin{align*}
\begin{aligned}
    &f_t(T_1,T_2,D_2,\lambda) :=\\
    &\quad \frac{1}{|D_2|} \sum_{(\bm{x},y) \in D_t} \ell(T_2(\bm{x}),y) + \frac{\lambda}{|D_s|} \sum_{(\bm{x},y) \in D_s \cap D_t} \delta(T_1,T_2,\bm{x})
\end{aligned}
\end{align*}
Let $\bar{T}_{2,t}$ denote the decision tree obtained by removing the subtree rooted at $t$ from $T_2$.
Then, the value of the loss function $f$ restricted to the new leaf node $t$ in $T_2$ can be computed as follows.
\begin{align*}
\begin{aligned}
    &\bar{f}_t(T_1,T_2,D_2,\lambda) :=\\
    &\quad \frac{1}{|D_2|} \sum_{(\bm{x},y) \in D_t} \ell(\bar{T}_{2,t}(\bm{x}),y) + \frac{\lambda}{|D_s|} \sum_{(\bm{x},y) \in D_s \cap D_t} \delta(T_1,\bar{T}_{2,t},\bm{x})
\end{aligned}
\end{align*}
Then, the cost $\alpha_t$ can be computed as follows.
\begin{align}
    \alpha_t = \frac{\bar{f}_t(T_1,T_2,D_2,\lambda) - f_t(T_1,T_2,D_2,\lambda)}{\#\mathrm{DescendantLeaves}(t) - 1}
\end{align}
Here, $\#\mathrm{DescendantLeaves}(t)$ is the number of leaves in the subtree rooted at $t$ on $T_2$.

CCP iteratively removes the subtree rooted at the node $t$ with the minimum cost and then recomputes the costs on the resulting decision tree.
It is known that the costs of the removed nodes are guaranteed to be monotonically non-decreasing, and these costs serve as candidate values of $\alpha$ (Algorithm~\ref{alg:alpha-path}).
When performing pruning for a given value of $\alpha$, it suffices to continue removing subtrees as long as the cost of the candidate node does not exceed $\alpha$ (Algorithm~\ref{alg:ccp}).
Notably, the behavior of CCP based on Eq.~\eqref{eq:ccp} differs slightly from that of the standard CCP, in that it may produce sibling nodes with identical prediction values.
While this can be regarded as a consequence of preserving backward compatibility in explanations as much as possible, we recommend removing such sibling nodes from the perspective of reducing redundancy in the decision tree (Algorithm~\ref{alg:siblings}).
In our experiments on regression tasks, prediction equivalence is defined as having an absolute difference no greater than $0.1\%$ of the observed range of target values.
Taken together, the complete decision tree update algorithm, which combines CART-BCTX with CCP using $K$-fold cross-validation, is summarized in Algorithm~\ref{alg:update-cv}.

\begin{algorithm}[t]
\caption{$\textrm{Compute-$\alpha$-Path}(T_2,D_2)$}
\label{alg:alpha-path}
\begin{algorithmic}[1]
    \STATE $\mathcal{A} \leftarrow \{0\}$
    \WHILE{The root node of $T_2$ is not a leaf node}
        \STATE Compute $\alpha_t$ for all the internal nodes of $T_2$ according to $D_2$
        \STATE $t^* \leftarrow \text{argmin}_t\{\alpha_t\}$
        \STATE $\mathcal{A} \leftarrow \mathcal{A} \cup \{\alpha_{t^*}\}$
        \STATE $T_2 \leftarrow \bar{T}_{2,t^*}$
    \ENDWHILE
    \RETURN $\mathcal{A}$
\end{algorithmic}
\end{algorithm}

\begin{algorithm}[t]
\caption{$\textrm{CostComplexityPruning}(T_2,D,\alpha)$}
\label{alg:ccp}
\begin{algorithmic}[1]
    \WHILE{The root node of $T_2$ is not a leaf node}
        \STATE Compute $\alpha_t$ for all the internal nodes of $T_2$ treating $D$ as $D_2$
        \STATE $t^* \leftarrow \text{argmin}_t\{\alpha_t\}$
        \IFLINE{$\alpha_{t^*} > \alpha$}{\textbf{break}}
        \STATE $T_2 \leftarrow \bar{T}_{2,t^*}$
    \ENDWHILE
    \RETURN $T_2$
\end{algorithmic}
\end{algorithm}

\begin{algorithm}[t]
\caption{$\textrm{PruneEquivalentSiblings}(T)$}
\label{alg:siblings}
\begin{algorithmic}[1]
    \STATE Let $t$ be the root node of $T$
    \IF{$t$ is a leaf node}
        \RETURN The tree with only the node $t$
    \ENDIF
    \STATE Let $T_L$ and $T_R$ be the left and right subtrees of $T$, respectively
    \STATE $T_L \leftarrow \textrm{PruneEquivalentSiblings}(T_L)$
    \STATE $T_R \leftarrow \textrm{PruneEquivalentSiblings}(T_R)$
    \STATE Let $t_L$ and $t_R$ be the left and right children of $t$, respectively
    \IF{$t_L$ and $t_R$ are leaf nodes}
        \IF{$\begin{cases}y_{t_L} = y_{t_R}& (\text{classification})\\ |y_{t_L} - y_{t_R}| \leq \tau & (\text{regression})\end{cases}$}
            \RETURN The tree with only the node $t$
        \ENDIF
    \ENDIF
    \RETURN The tree rooted at $t$ with the left subtree $T_L$ and the right subtree $T_R$
\end{algorithmic}
\end{algorithm}

\begin{algorithm}[t]
\caption{$\textrm{UpdateTree-CV}(T_1,D_2,\lambda,K)$}
\label{alg:update-cv}
\begin{algorithmic}[1]
    \STATE $T_2 \leftarrow \textrm{CART-BCTX}(D_2,(x_j^{\mathrm{min}})_{j=1}^m,(x_j^{\mathrm{max}})_{j=1}^m,0)$
    \STATE $\mathcal{A} \leftarrow \textrm{Compute-$\alpha$-Path}(T_2,D_2)$
    \STATE $f_\alpha \leftarrow 0$ for all $\alpha \in \mathcal{A}$
    \FOR{$k=1,\ldots,K$}
        \STATE $D_2^{\mathrm{trn}} \leftarrow \text{Get training set of $k$-th fold over}~D_2$
        \STATE $D_2^{\mathrm{val}} \leftarrow \text{Get validation set of $k$-th fold over}~D_2$
        \STATE $T_2^k \leftarrow \textrm{CART-BCTX}(D_2^{\mathrm{trn}},(x_j^{\mathrm{min}})_{j=1}^m,(x_j^{\mathrm{max}})_{j=1}^m,0)$
        \FOR{$\alpha \in \mathcal{A}$ in ascending order}
            \STATE $T_2^k \leftarrow \textrm{CostComplexityPruning}(T_2^k,D_2^{\mathrm{trn}},\alpha)$
            \STATE $T_2^{k,\alpha} \leftarrow \textrm{PruneEquivalentSiblings}(T_2^k)$
            \STATE $f_\alpha \leftarrow f_\alpha + f(T_1,T_2^{k,\alpha},D_2^{\mathrm{val}},\lambda)$
        \ENDFOR
    \ENDFOR
    \STATE $\alpha^* \leftarrow \text{argmin}_\alpha\{f_\alpha\}$
    \STATE $T_2^\mathrm{ccp} \leftarrow \textrm{CostComplexityPruning}(T_2,D_2,\alpha^*)$
    \STATE $T_2^\mathrm{ccp} \leftarrow \textrm{PruneEquivalentSiblings}(T_2^\mathrm{ccp})$
    \RETURN $T_2^\mathrm{ccp}$
\end{algorithmic}
\end{algorithm}

\section{Additional Experiments under Alternative Update Settings}
In the main body, we assumed that the old dataset $D_1$ is contained in the new dataset $D_2$.
However, depending on data retention policies in practical settings, two additional update scenarios can be considered.
One scenario is where $D_1$ and $D_2$ partially overlap, and the other is where $D_1$ and $D_2$ are disjoint.
In the following, we present additional experimental results under these two settings.
The datasets and the configurations of all algorithms are the same as those used in the main experiments.

\subsection{Partially Overlapping Old and New Datasets}
\begin{figure*}[t]
    \centering
    \includegraphics[width=\textwidth]{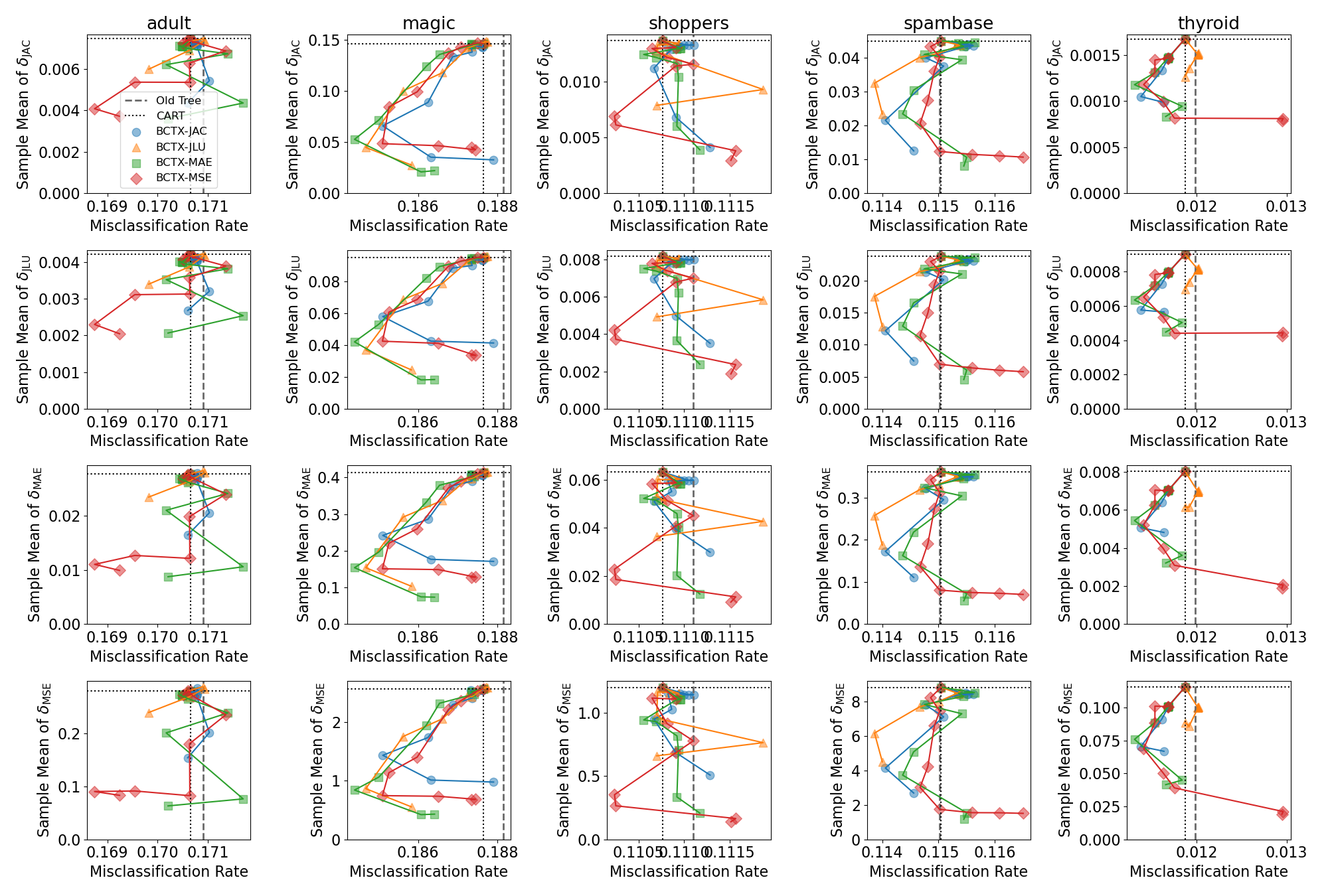}
    \caption{
        Trade-offs for classification tasks under partially overlapping old and new datasets.
    }
    \Description{Results for classification tasks under partially overlapping old and new datasets.}
    \label{fig:classif-overlap}
\end{figure*}

\begin{figure*}[t]
    \centering
    \includegraphics[width=\textwidth]{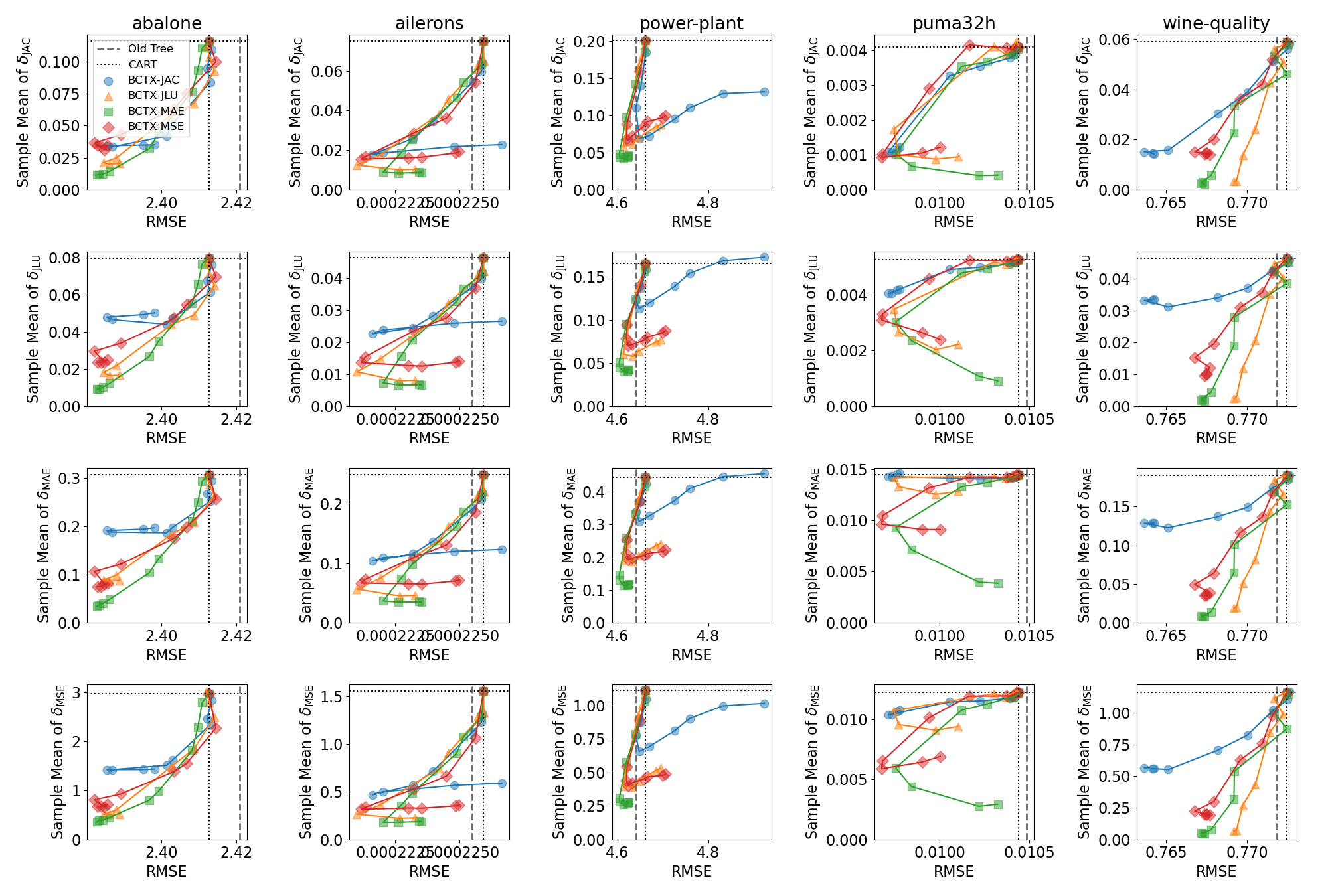}
    \caption{
        Trade-offs for regression tasks under partially overlapping old and new datasets.
    }
    \Description{Results for regression tasks under partially overlapping old and new datasets.}
    \label{fig:regress-overlap}
\end{figure*}

\begin{figure*}[t]
    \centering
    \includegraphics[width=\textwidth]{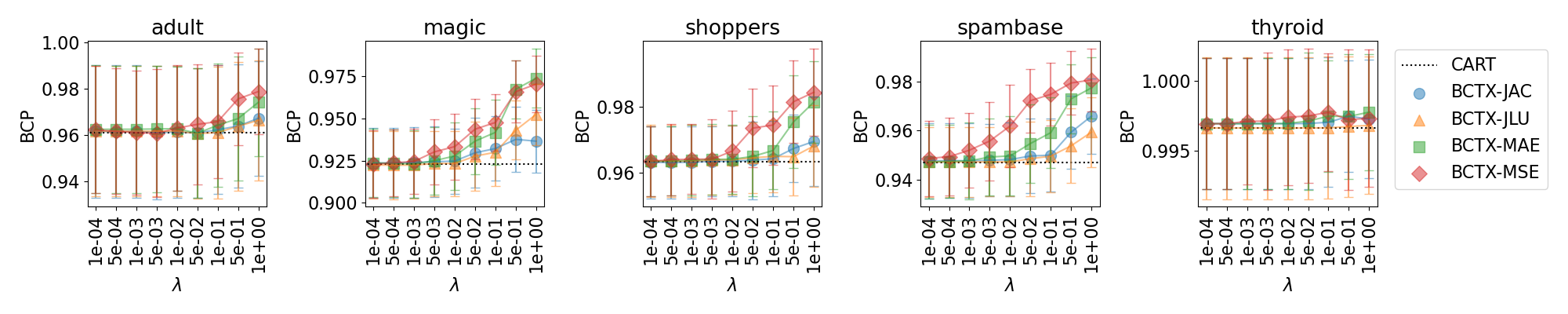}
    \centering
    \includegraphics[width=\textwidth]{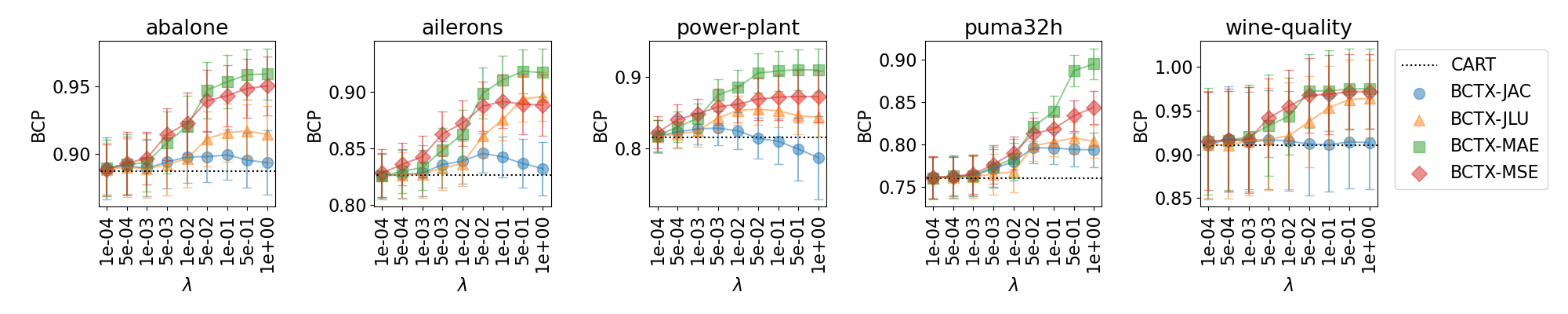}
    \caption{
        BCP values (higher is better) for CART and CART-BCTX under partially overlapping old and new datasets.
    }
    \Description{Results for BCP under partially overlapping old and new datasets.}
    \label{fig:bcpr-overlap}
\end{figure*}

\begin{figure*}[t]
    \centering
    \includegraphics[width=\textwidth]{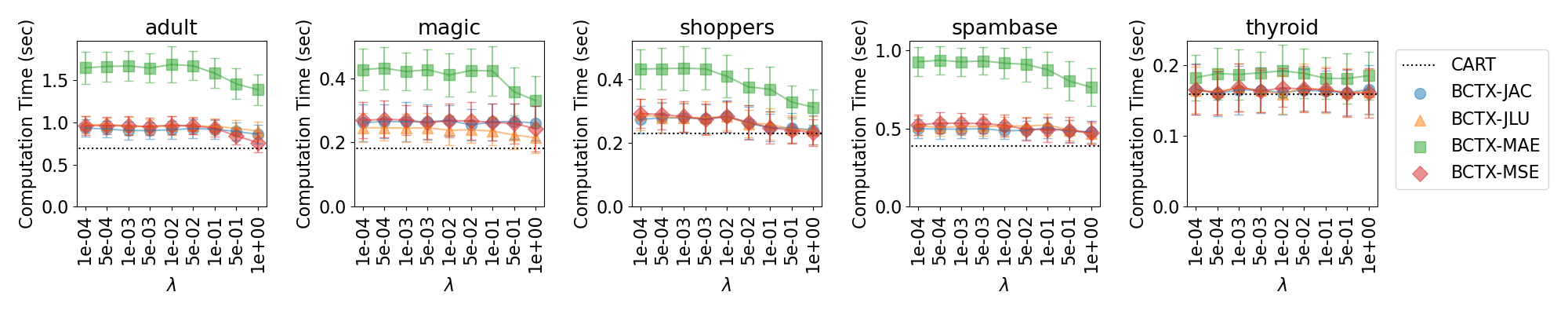}
    \centering
    \includegraphics[width=\textwidth]{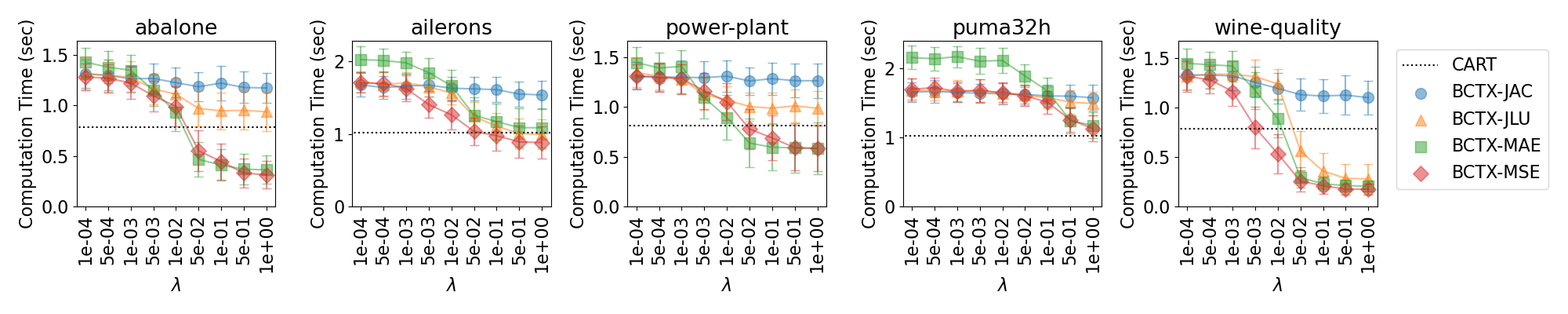}
    \caption{
        Computation times for CART and CART-BCTX under partially overlapping old and new datasets.
    }
    \Description{Results for computation times under partially overlapping old and new datasets.}
    \label{fig:time-overlap}
\end{figure*}

We set the number of samples in the old dataset $D_1$ to $1,000$ and that in the new dataset $D_2$ to $1,000$.
In addition, we assume the number of common samples in $D_1 \cap D_2$ to $500$.
We conducted $100$ random trials for each dataset.
Figure~\ref{fig:classif-overlap} (for classification tasks) and Figure~\ref{fig:regress-overlap} (for regression tasks) present the predictive performance and BCLTX values of the updated decision trees.
Figure~\ref{fig:bcpr-overlap} shows the BCP baseline obtained by CART and the BCP changes achieved by CART-BCTX with varying $\lambda$.
Figure~\ref{fig:time-overlap} presents the computation times of CART and CART-BCTX.

Although the trajectories with varying $\lambda$ are somewhat complex, CART-BCTX generally exhibits a favorable trade-offs between predictive performances and BCLTX values.
In terms of predictive performance, naive updates using CART may in some cases degrade performance relative to the old decision tree, whereas CART-BCTX achieves substantial improvements in most cases.
This behavior can be attributed to the fact that CART-BCTX learns a new decision tree while explicitly referencing the old decision tree.
For both the trade-off analysis and BCP changes, improvements obtained using $\delta_{\mathrm{MAE}}$ or $\delta_{\mathrm{MSE}}$ are particularly effective, and execution of CART-BCTX with $\delta_{\mathrm{MAE}}$ or $\delta_{\mathrm{MSE}}$ is recommended under this update setting as well.
In addition, the computation times of CART and CART-BCTX remain comparable.

\subsection{Disjoint Old and New Datasets}
\begin{figure*}[t]
    \centering
    \includegraphics[width=\textwidth]{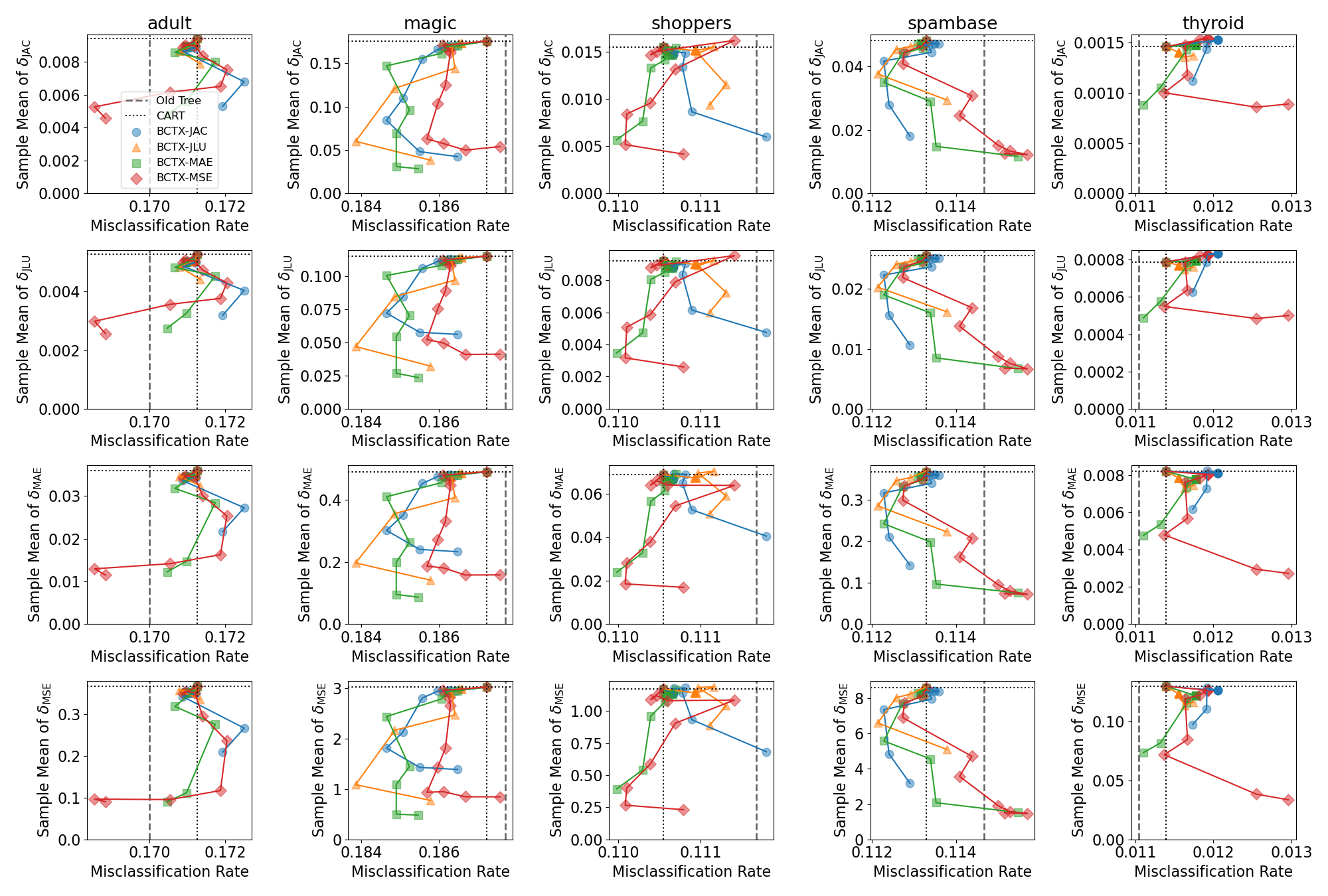}
    \caption{
        Trade-offs for classification tasks under disjoint old and new datasets.
    }
    \Description{Results for classification tasks under disjoint old and new datasets.}
    \label{fig:classif-disjoint}
\end{figure*}

\begin{figure*}[t]
    \centering
    \includegraphics[width=\textwidth]{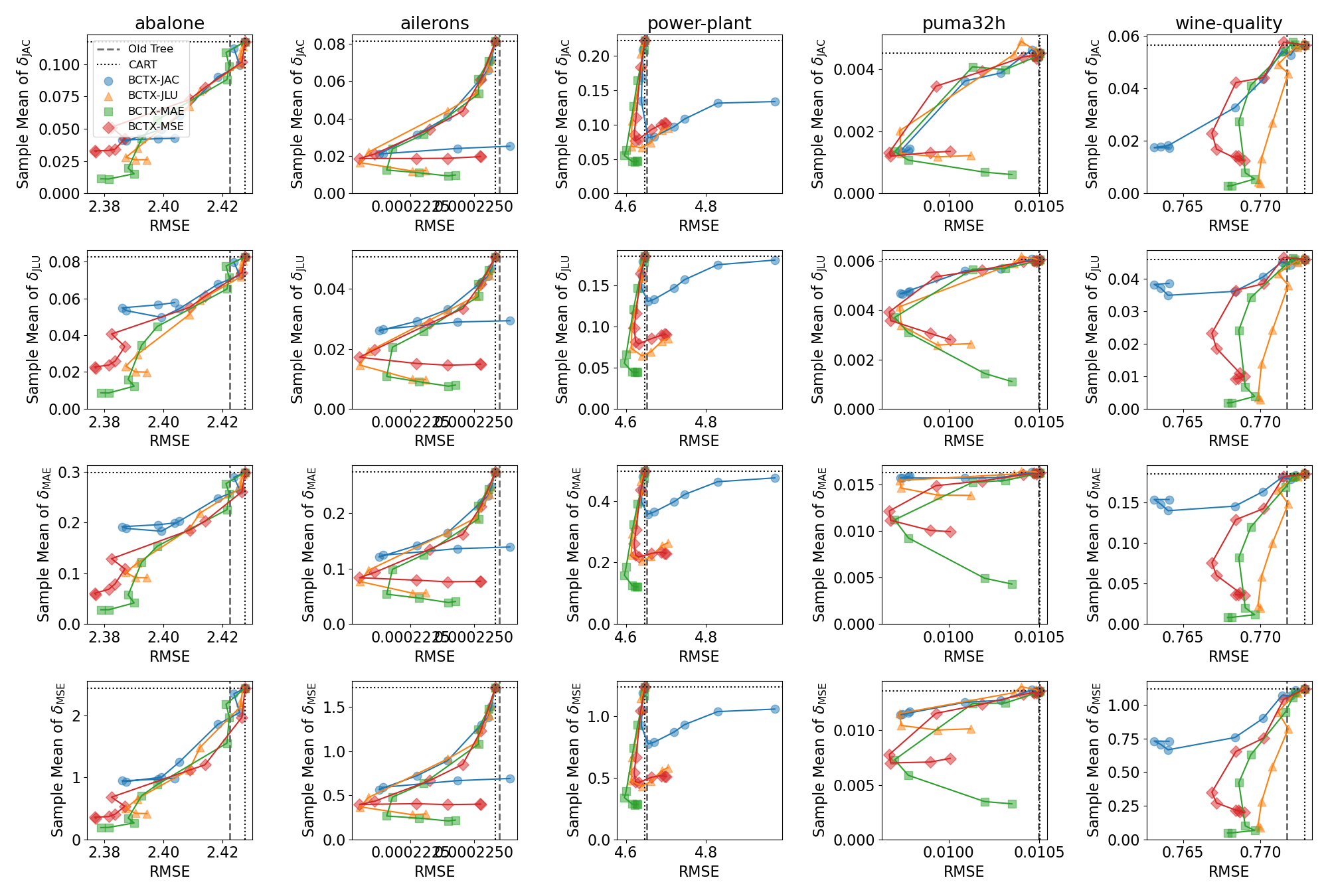}
    \caption{
        Trade-offs for regression tasks under disjoint old and new datasets.
    }
    \Description{Results for regression tasks under disjoint old and new datasets.}
    \label{fig:regress-disjoint}
\end{figure*}

\begin{figure*}[t]
    \centering
    \includegraphics[width=\textwidth]{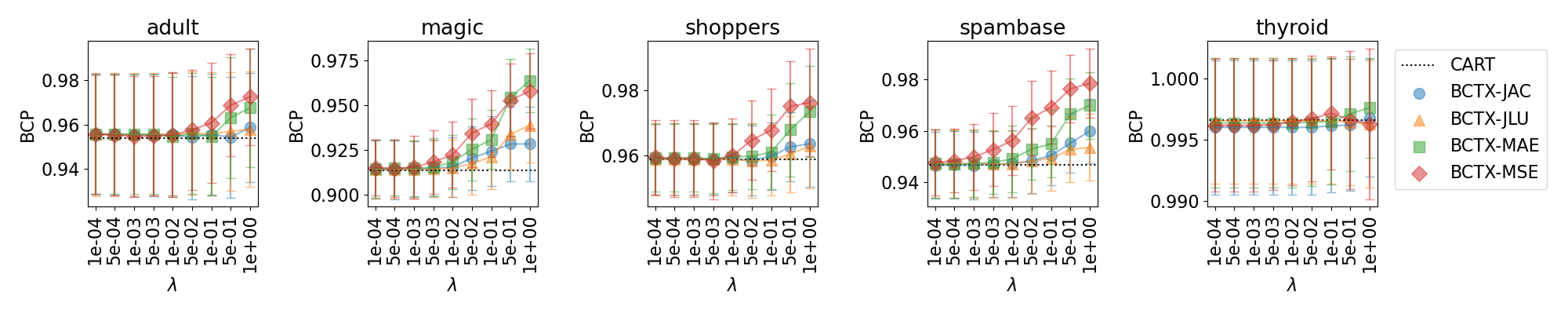}
    \centering
    \includegraphics[width=\textwidth]{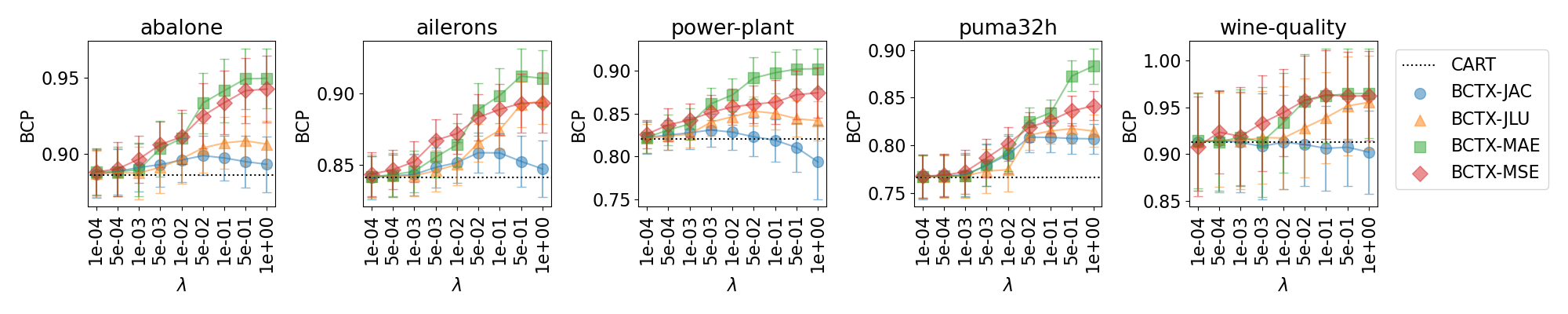}
    \caption{
        BCP values (higher is better) for CART and CART-BCTX under disjoint old and new datasets.
    }
    \Description{Results for BCP under disjoint old and new datasets.}
    \label{fig:bcpr-disjoint}
\end{figure*}

\begin{figure*}[t]
    \centering
    \includegraphics[width=\textwidth]{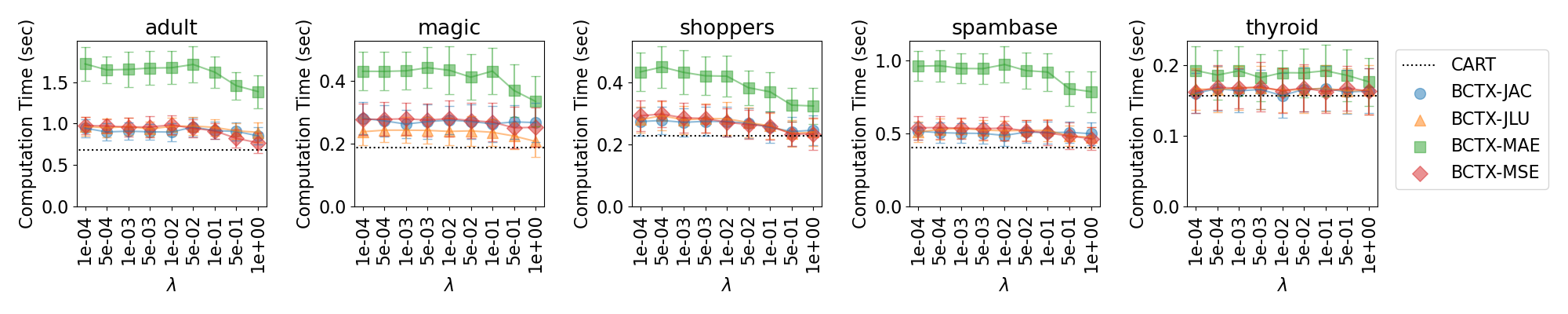}
    \centering
    \includegraphics[width=\textwidth]{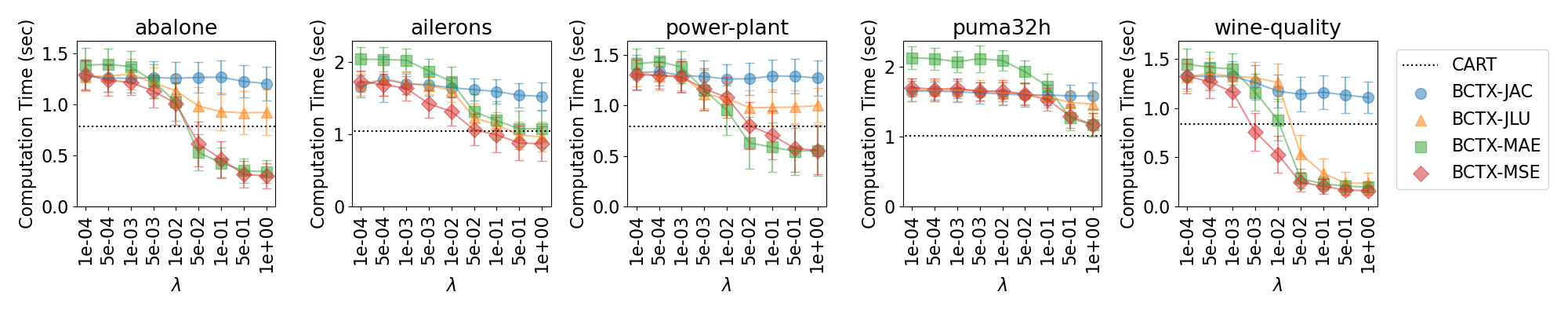}
    \caption{
        Computation times for CART and CART-BCTX under disjoint old and new datasets.
    }
    \Description{Results for computation times under disjoint old and new datasets.}
    \label{fig:time-disjoint}
\end{figure*}

We set the number of samples in the old dataset $D_1$ to $1,000$ and that in the new dataset $D_2$ to $1,000$.
In addition, we assume $D_1 \cap D_2 = \emptyset$.
We conducted $100$ random trials for each dataset.
Figure~\ref{fig:classif-disjoint} (for classification tasks) and Figure~\ref{fig:regress-disjoint} (for regression tasks) present the predictive performance and BCLTX values of the updated decision trees.
Figure~\ref{fig:bcpr-disjoint} shows the BCP baseline obtained by CART and the BCP changes achieved by CART-BCTX with varying $\lambda$.
Figure~\ref{fig:time-disjoint} presents the computation times of CART and CART-BCTX.

Except for the `adult` and `thyroid` datasets, CART-BCTX generally exhibits favorable trade-offs between predictive performances and BCLTX values.
For the `adult` dataset, substantial improvements are observed with $\delta_{\mathrm{MSE}}$ for large $\lambda$.
For the `thyroid` dataset, moderate improvements are observed with $\delta_{\mathrm{MAE}}$ for large $\lambda$.
For both the trade-off analysis and BCP changes, improvements obtained using $\delta_{\mathrm{MAE}}$ or $\delta_{\mathrm{MSE}}$ are again effective, and execution of CART-BCTX with $\delta_{\mathrm{MAE}}$ or $\delta_{\mathrm{MSE}}$ is recommended under this update setting as well.
In addition, the computation times of CART and CART-BCTX remain comparable.

\section{Visualized Tree Comparison}
\label{sec:appendix-visualized-tree-comparizon}
\begin{figure*}[t]
    % CART
    \subfigure[CART (Old)]{\includegraphics[width=0.34\textwidth]{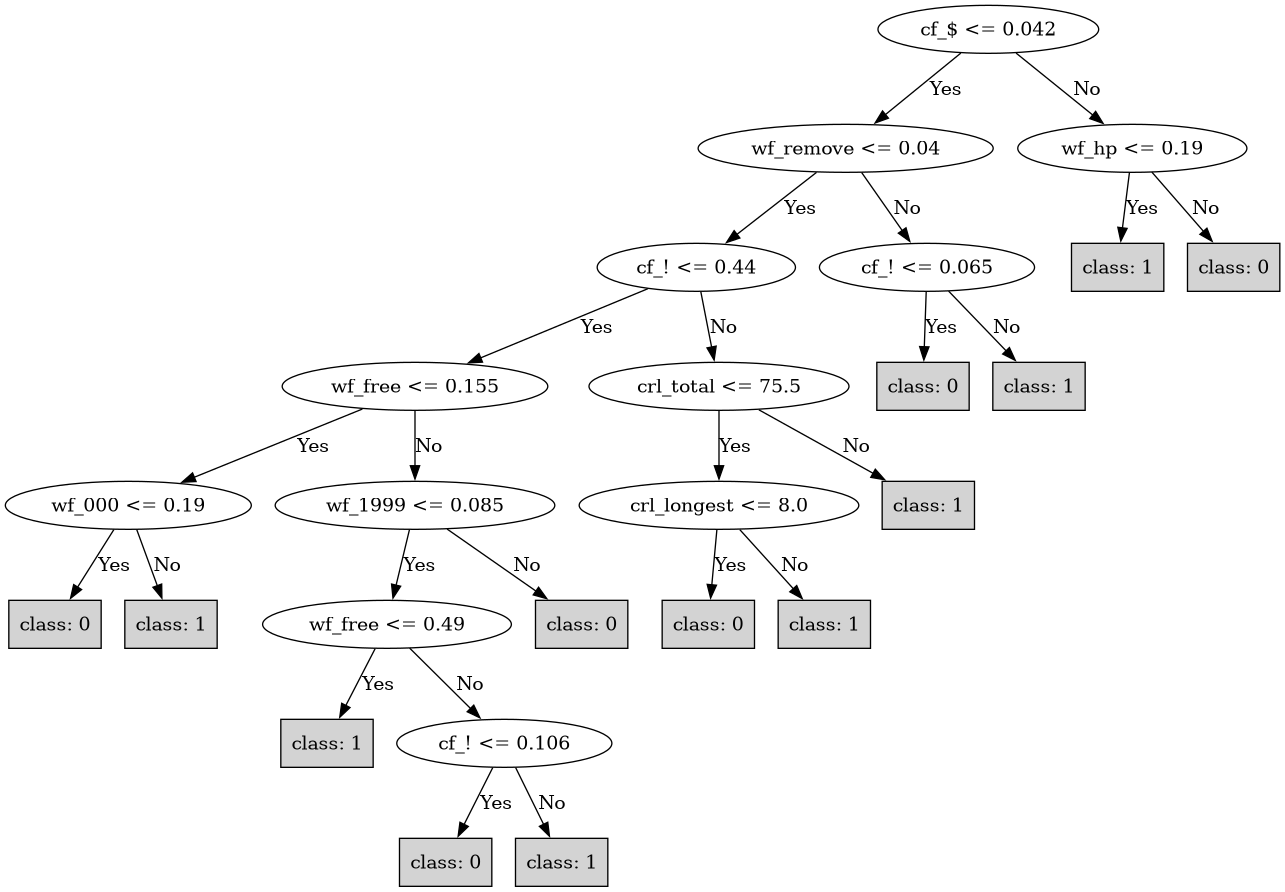}}
    \subfigure[CART (New, Naive)]{\includegraphics[width=0.34\textwidth]{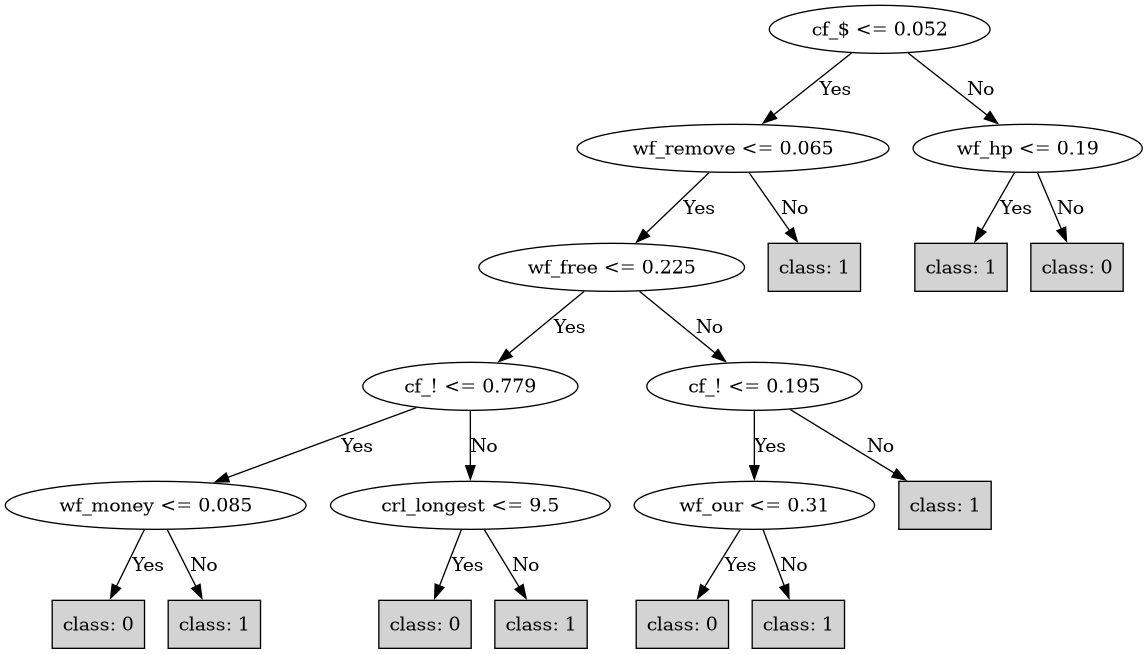}}
    \subfigure[CART (New, BCTX)]{\includegraphics[width=0.3\textwidth]{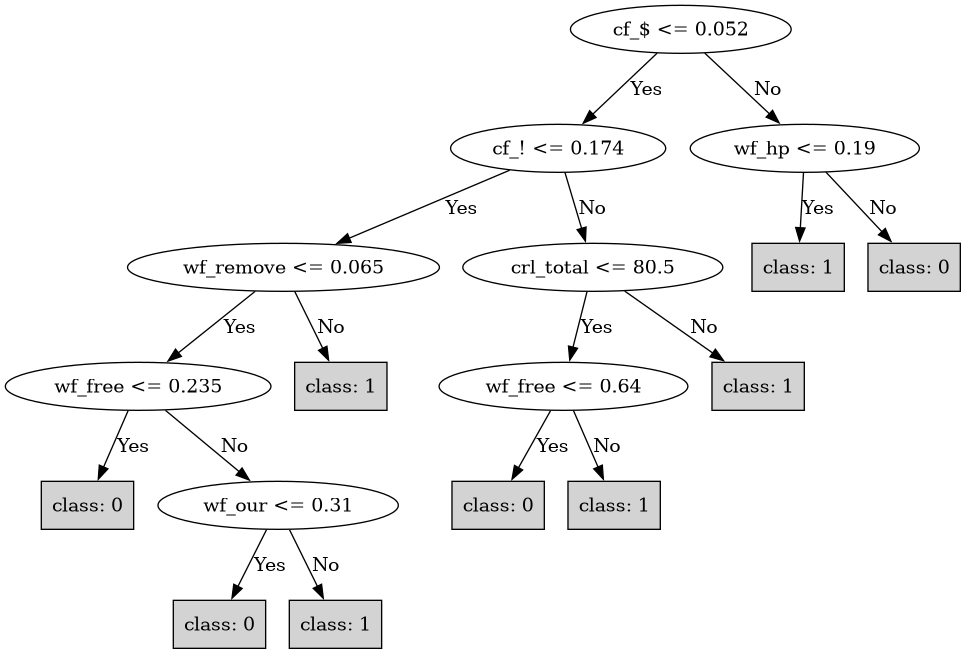}}
    \\
    % VFDT
    \subfigure[VFDT (Old)]{\includegraphics[width=0.17\textwidth]{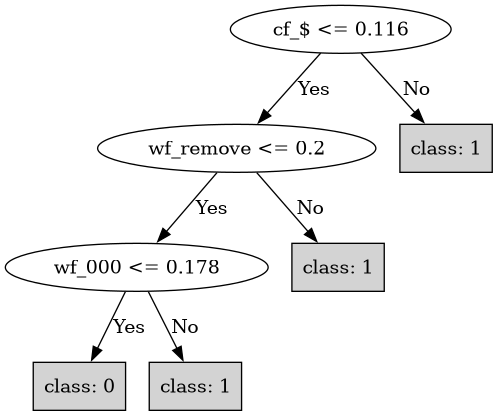}}
    \subfigure[VFDT (New)]{\includegraphics[width=0.26\textwidth]{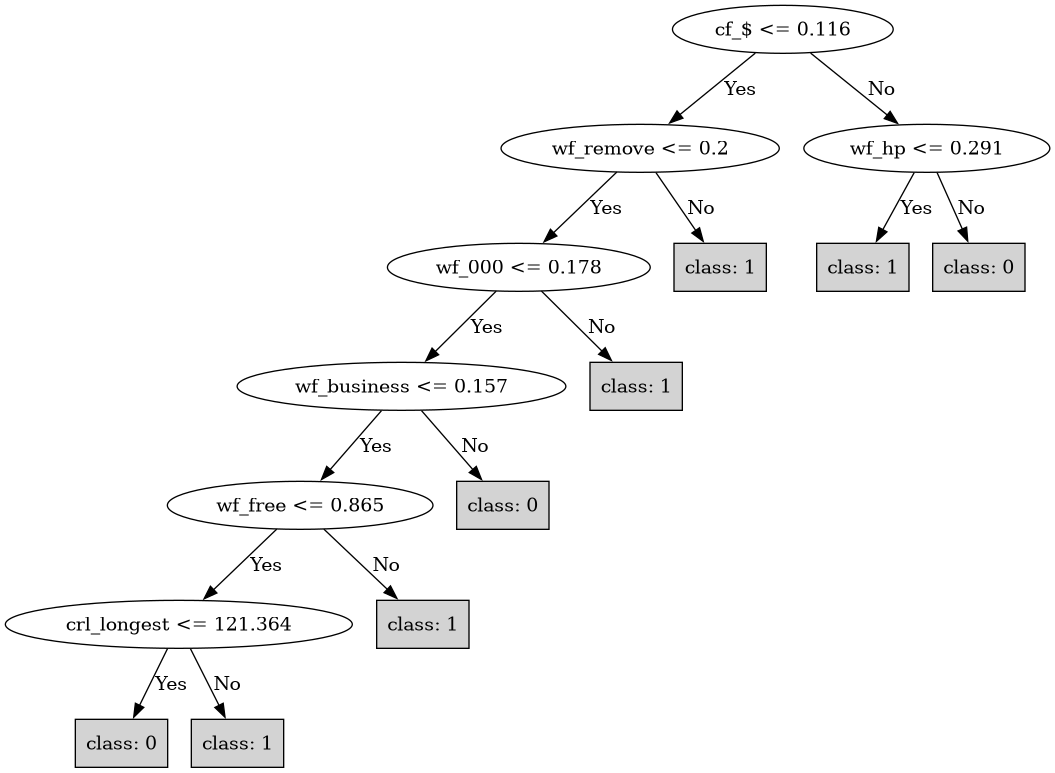}}
    % HAT
    \subfigure[HAT (Old)]{\includegraphics[width=0.19\textwidth]{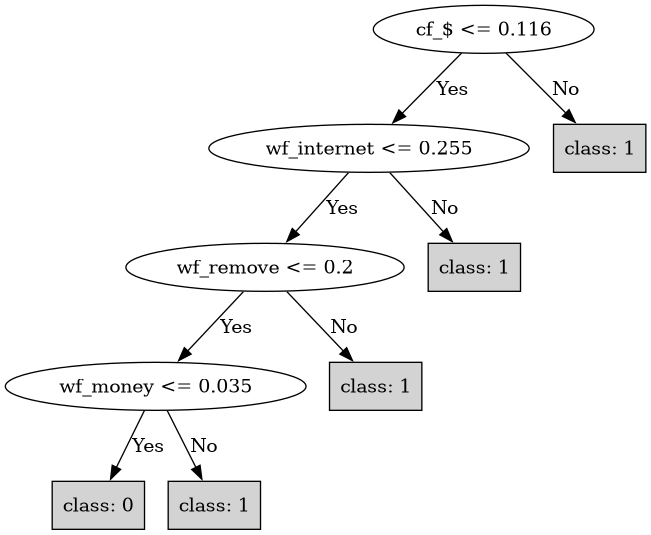}}
    \subfigure[HAT (New)]{\includegraphics[width=0.36\textwidth]{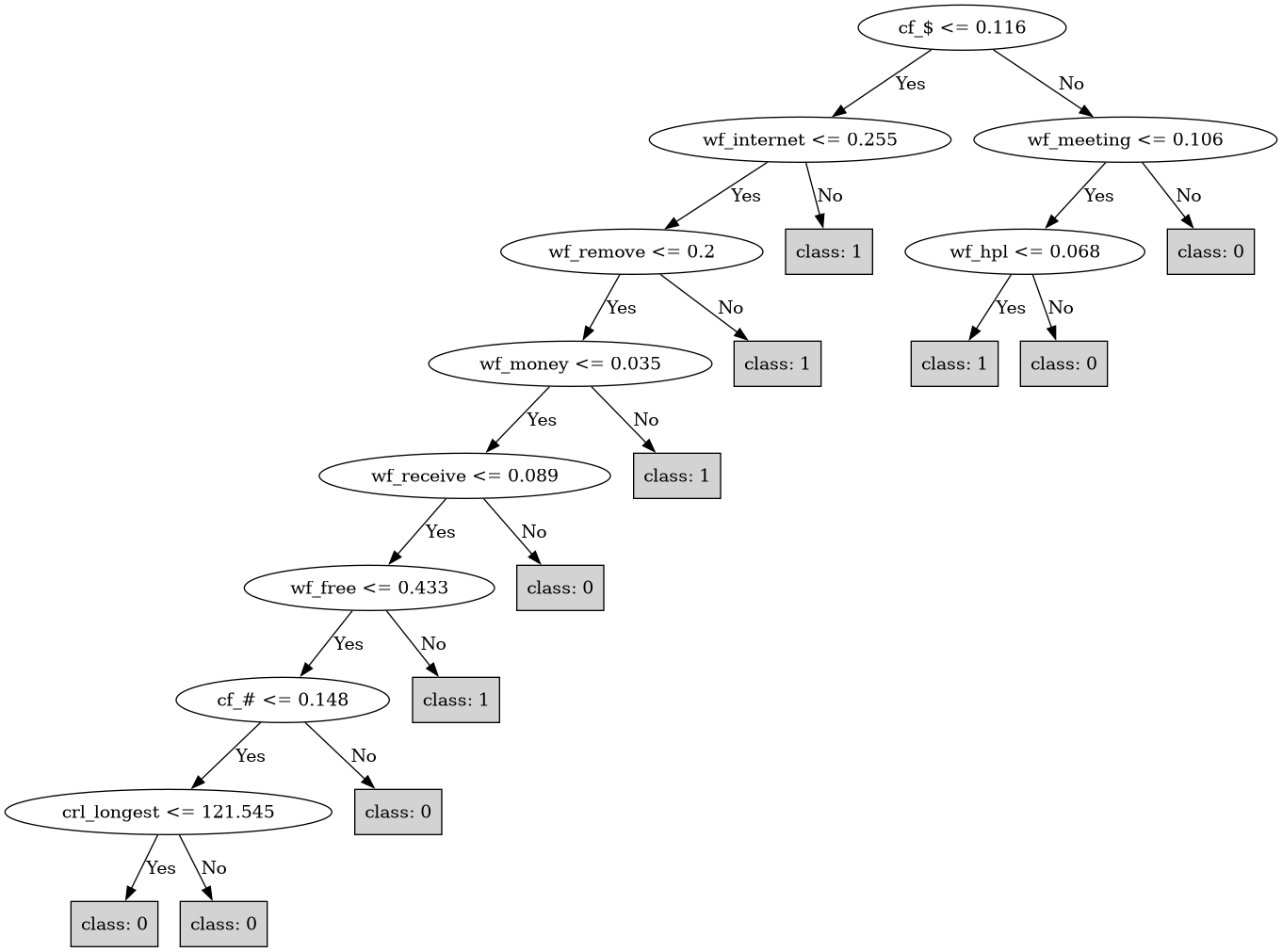}}
    % caption
    \caption{
        Specific examples of old/new decision trees on the spambase dataset.
    }
    \Description{Visualized Tree Comparison.}
    \label{fig:tree-comparison}
\end{figure*}

In this section, we compare specific examples of old/new decision trees, on the spambase dataset, obtained using each method (CART, CART-BCTX, VFDT, and HAT) in Section~\ref{sec:comparison-with-incremental-trees}.
The trees are illustrated in Figure~\ref{fig:tree-comparison}.
The task of spambase dataset is to determine whether a given email is spam (class 1) or not (class 0) via word/character frequencies and capital run lengthes.
In the tree illustrations, we use the abbreviations ``wf'' for ``word\_freq'', ``cf'' for ``char\_freq'', and ``crl'' for ``capital\_run\_length''.

First, we compare the new decision trees (b) and (c) of CART and CART-BCTX with the old decision tree (a) of CART.
Although there was only a little difference in the features added or removed by the updates (a)$\rightarrow$(b) and (a)$\rightarrow$(c), the update (a)$\rightarrow$(c) made smaller changes in the thresholds of the shared features than the update (a)$\rightarrow$(b).
As a result, the changes in the decision boundaries were smaller for the update (a)$\rightarrow$(c).
In fact, $\delta_\mathrm{MSE}$ was $4.17$ and $2.35$ for the updates (a)$\rightarrow$(b) and (a)$\rightarrow$(c), respectively.
Thus, \emph{we could visually confirm the effectiveness of CART-BCTX for the backward compatibility in tree-based explanations}.

Next, we focus on the results of incremental decision trees VFDT and HAT.
The new decision trees (e) and (g) are generated by adding some new branches to the old ones (d) and (f).
Furthermore, most of these new branches are based on features that are not used in the old trees.
As a result, both the updates (d)$\rightarrow$(e) and (f)$\rightarrow$(g) significantly changed the decision boundaries.
In fact, the values of $\delta_\mathrm{MSE}$ were $13.23$ for (d)$\rightarrow$(e) and $18.83$ for (f)$\rightarrow$(g), respectively.
Thus, \emph{HAT may not necessarily maintain backward compatibility in tree-based explanations}.

\section{A Preliminary Analysis of CART-BCTX under Concept Drift Settings}

\label{sec:under-concept-drift}
In this section, we present a preliminary analysis of how well CART-BCTX preserves predictive performance and backward compatibility when a decision tree is repeatedly updated in a streaming environment with concept drift.
We update and evaluate the tree using a "test-then-train" approach for sequentially arriving batches, and quantify the prediction loss, backward compatibility in tree-based explanations, and backward compatibility in predictions before and after each update.

\subsection{Setup}
\paragraph{Dataset}
We used the elec2 dataset~\cite{elec2} (45,312 samples in total), which consists of real-world data and is available in the River library~\cite{montiel2021river}.
The task is a binary classification regarding electricity supply and demand (whether the price is UP or DOWN).
The features included the continuous variables ``nswprice'', ``nswdemand'', ``vicprice'', ``vicdemand'', and ``transfer'', as well as the ``day'' of the week (one-hot encoded as 7 categories) and the time period, which was transformed using $\sin$ and $\cos$ functions to preserve periodicity: ``period\_sin'' and ``period\_cos''.
The stream was processed in batch-wise manner by dividing the chronologically ordered samples into batches at fixed intervals.
Taking advantage of the fact that there are 48 observations per day (at 30-minute intervals), we configured three conditions with different update frequencies by setting the batch length to 1-week, 2-week, and 4-week ($48 \times 7$ samples per week).
The first batch was used for initial training, and evaluation and updates were performed on each subsequent batch.

\paragraph{Competitors}
We compared CART-BCTX, CART, HAT, and a decision tree without update (the CART baseline for prediction loss).
For HAT, we used the implementation from the River library and set ``grace\_period=48'', ``delta=1e-5'', and ``tau=0.01'' for accommodating the periodicity of the elec2 dataset.
For CART-BCTX, we set $\lambda = 0.005$ and $\delta = \delta_\mathrm{MSE}$.
For both CART and CART-BCTX, no depth limit was imposed, the minimum sample size was set to 1\% of the number of samples in the training batch, and the post-pruning process by CCP was performed using 5-fold cross-validation.

\begin{figure*}[t]
    \centering
    \includegraphics[width=\textwidth]{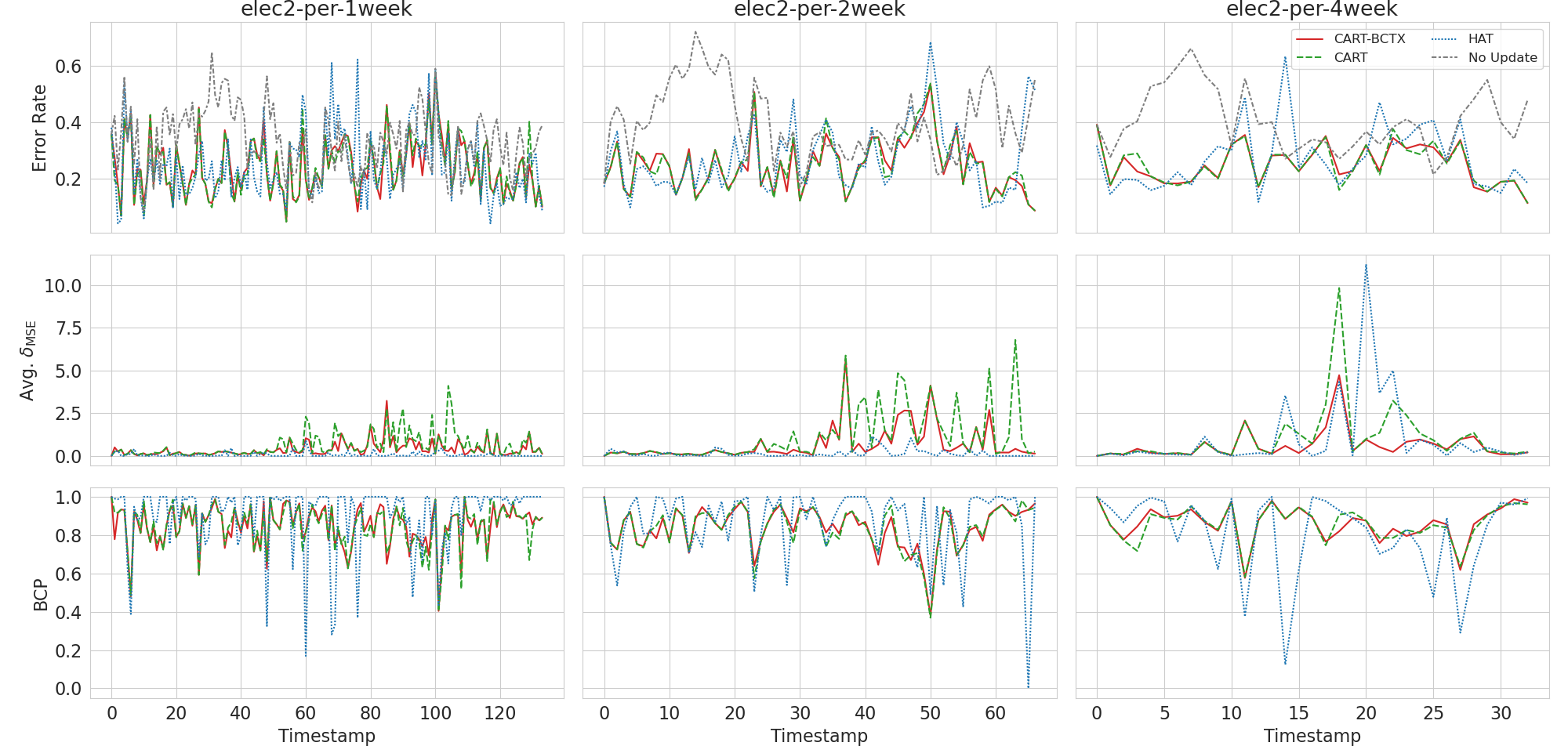}
    % caption
    \caption{
        Performance under concept drift situation using elec2 dataset.
    }
    \Description{Performance under concept drift.}
    \label{fig:elec2}
\end{figure*}

\subsection{Results}
Figure~\ref{fig:elec2} summarizes the error rates, sample-averaged BCLTX ($\delta_\mathrm{MSE}$), and BCP values of CART, CART-BCTX, and HAT.
We evaluate these metrics across sequential batches with different update intervals (1-week, 2-week, and 4-week settings), reporting only the error rate for the no-update baseline.

First, in terms of predictive performance, CART and CART-BCTX consistently achieved stable and superior error rates across all settings. 
In contrast, HAT occasionally showed noticeable degradation, sometimes performing worse than the no-update baseline. 
This indicates that batch-wise retraining remains robust under repeated updates, whereas HAT may exhibit instability.

Second, regarding BCLTX, HAT consistently achieved low values in the 1-week and 2-week settings, indicating strong consistency with recently observed data (336 and 672 samples). 
However, in the 4-week setting (1,334 samples), HAT occasionally showed clear increases in BCLTX, suggesting degradation from a medium- to long-term periods.
In contrast, CART exhibited large fluctuations, while CART-BCTX reduced these fluctuations and maintained more stable values for all settings.

Third, in terms of BCP, HAT occasionally exhibited notable drops in several batches, indicating instability in preserving previously correct predictions. 
In contrast, CART and CART-BCTX showed stable BCP trends with comparable performance across all settings. 
This suggests that enforcing backward compatibility in tree-based explanations does not harm prediction consistency.

In summary, \emph{CART-BCTX achieves a balanced trade-off between predictive performance and backward compatibility, with more stable behavior than CART and more consistent medium- to long-term periods than HAT}.

\end{document}